\documentclass{article}

\PassOptionsToPackage{numbers, compress}{natbib}
\usepackage[final]{neurips_2023}
\usepackage[utf8]{inputenc} 
\usepackage[T1]{fontenc}    
\usepackage{hyperref}       
\usepackage{url}            
\usepackage{booktabs}       
\usepackage{amsfonts}       
\usepackage{nicefrac}       
\usepackage{microtype}      
\usepackage{xcolor}         

\usepackage{float}
\usepackage{bbold}
\usepackage{amssymb}
\usepackage{amsmath}
\usepackage{amsthm}
\usepackage{amsfonts}
\newtheorem{theorem}{Theorem}

\newtheorem{proposition}[theorem]{Proposition}

\usepackage[pdftex]{graphicx}
\usepackage{natbib}
\title{Addressing the speed-accuracy simulation trade-off for adaptive spiking neurons}

\author{%
  Luke Taylor\\
  Department of Physiology, Anatomy and Genetics\\
  University of Oxford\\
  Oxford, United Kingdom \\
  \texttt{luke.taylor@hertford.ox.ac.uk} \\
  \And
  Andrew J King \\
  Department of Physiology, Anatomy and Genetics\\
  University of Oxford\\
  Oxford, United Kingdom \\
  \texttt{andrew.king@dpag.ox.ac.uk} \\
  \AND
  Nicol S Harper \\
  Department of Physiology, Anatomy and Genetics\\
  University of Oxford\\
  Oxford, United Kingdom \\
  \texttt{nicol.harper@dpag.ox.ac.uk} \\
}

\begin{document}

\maketitle

\begin{abstract}
The adaptive leaky integrate-and-fire (ALIF) model is fundamental within computational neuroscience and has been instrumental in studying our brains \textit{in silico}. Due to the sequential nature of simulating these neural models, a commonly faced issue is the speed-accuracy trade-off: either accurately simulate a neuron using a small discretisation time-step (DT), which is slow, or more quickly simulate a neuron using a larger DT and incur a loss in simulation accuracy. Here we provide a solution to this dilemma, by algorithmically reinterpreting the ALIF model, reducing the sequential simulation complexity and permitting a more efficient parallelisation on GPUs. We computationally validate our implementation to obtain over a $50\times$ training speedup using small DTs on synthetic benchmarks. We also obtained a comparable performance to the standard ALIF implementation on different supervised classification tasks - yet in a fraction of the training time. Lastly, we showcase how our model makes it possible to quickly and accurately fit real electrophysiological recordings of cortical neurons, where very fine sub-millisecond DTs are crucial for capturing exact spike timing.
\end{abstract}

\section{Introduction}
The surge of progress in artificial neural networks (ANNs) over the last decade has advanced our understanding of the potential computational principles underlying the processing of sensory information \cite{harper2016network,  singer2023hierarchical, cadena2019deep, francl2022deep, yamins2016using, bakhtiari2021functional, mineault2021your, ocko2018emergence, conwell2021neural}. Although these networks architecturally bear a resemblance to the brain \cite{richards2019deep}, they tend to omit a key physiological constraint: the spike. With their increased biological realism, spiking neural networks (SNNs) have shown great promise in bridging the gap between experimental data and computational models. SNNs can be fitted to real neural data \cite{jolivet2008quantitative, kobayashi2009made, rossant2011fitting, mensi2012parameter, pozzorini2013temporal}, or used to simulate the brain, offering a new level of understanding of the complex workings of the nervous system \cite{deneve2016efficient, vogels2011inhibitory, confavreux2020meta, braun2021online}. They also have engineering applications in energy-efficient machine learning \cite{wunderlich2019demonstrating}. 

\begin{figure}
    \includegraphics[width=1\linewidth]{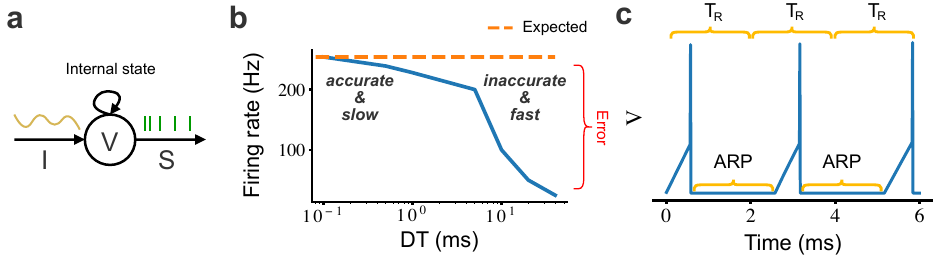}
	\centering
	\caption{\textbf{Problem overview.} \textbf{a.} Schematic of an ALIF neuron: input current $I$ charges membrane potential $V$ and outputs spikes $S$ if firing threshold is reached (with the neuron's internal state evolving over time). \textbf{b.} An example of the simulation trade-off problem when simulating a single ALIF neuron with fixed synaptic weights receiving Poisson spike input. The simulation error and the speed grow for increasing discretisation time (DT). \textbf{c.} Observation for our solution: a neuron emits at most a single spike during a simulation span $T_R$ equal in length to the neuron's absolute refractory period (ARP).}
 	\label{fig:figure1}
\end{figure}

An established class of spiking models in computational neuroscience is the leaky integrate-and-fire (LIF) neuron, with origins dating back to 1907 \cite{lapicque1907recherches}. Just like in real neurons, input current (resulting from presynaptic input) charges the membrane potential of the model neurons, which then output binary signals (\textit{i.e.} spikes) as a form of communication (Figure \ref{fig:figure1}a). The adaptive leaky integrate-and-fire (ALIF) model \cite{gerstner2014neuronal} is a modern extension of the LIF. It more closely mimics the biology, capturing a key property of neurons, which is their adaptive firing threshold (\textit{i.e.} spikes become less frequent in response to a steady input current \cite{kandel2000principles}). ALIF neurons have been shown to accurately fit real neural recordings \cite{jolivet2008quantitative, gerstner2014neuronal, levakova2019adaptive, zeng2021temporal} and to outperform the simpler LIF neurons on various machine learning benchmarks \cite{bellec2018long, bellec2020solution, yin2020effective, yin2021accurate}.

Despite these modelling advances, a major shortcoming of LIF and ALIF neurons is their slow inference and training times. Unlike real neurons, modelling neuron dynamics involves sequential computation over discretised time. This leads to a problematic trade-off between speed and accuracy when simulating SNNs. A small DT enables accurate modelling of dynamics, but is slow to stimulate and train on computer systems such as GPUs. A large DT obtains less accurate dynamics, but at the benefit of being faster to simulate and train \cite{perezinit} (Figure \ref{fig:figure1}b). This raises the important question of capturing the best of both worlds: \textbf{is there a way to accelerate the inference and training of spiking LIF and ALIF neurons without sacrificing simulation accuracy?}

In this work, we address the speed-accuracy trade-off when simulating and training LIF and ALIF SNNs, and present a solution that is both fast and accurate. We take advantage of a fundamental property of neurons that is sometimes not modelled - the absolute refractory period (ARP). This is a short period of time following spike initiation, during which a neuron cannot fire again. As a result, a neuron can spike at most once within such a period (Figure \ref{fig:figure1}c). We leverage this observation to develop a novel algorithmic reformulation of the ALIF model which reduces the sequential simulation complexity of the standard ALIF algorithm. Specifically, we outline how ALIF recurrent networks can be simulated with a constant sequential complexity $O(1)$ over the ARP simulation length $T_R$, and how this approach can be extended to longer simulation lengths to obtain identical dynamics to the standard ALIF network algorithm. Faster simulation and training are theoretically obtained for growing length $T_R$, by either employing physiologically plausible ARPs of $\sim 2$ms and decreasing the DT to a very fine value $\sim 0.1$ms (for realistic neural modelling), or setting the DT to a coarser value and adopting larger non-physiological ARPs (for machine learning tasks). Our main contributions are the following:
\begin{itemize}
	\item We develop a novel algorithmic reformulation of the ALIF model with an $O(T/T_R)$ - rather than $O(T)$ - sequential complexity, for simulation length $T$ and ARP length $T_R$.\footnote{To avoid ambiguity, simulation length $T$ and ARP length $T_R$ are number of time steps (dimensionless) and not unit time.}
	\item We find that our model achieves substantial inference (up to $40\times$) and training speedup (up to $53\times$) over the standard ALIF SNN for increasing ARP time steps, and find this to hold over different numbers of simulation lengths, batch sizes, number of neurons and layers.
	\item We demonstrate the feasibility of our model to be trained using surrogate gradient descent, with our accelerated ALIF SNN achieving comparable accuracies to the standard ALIF SNN on temporal spiking classification datasets - yet in a fraction of the training time.
	\item Finally, we showcase how our ALIF implementation makes it possible to quickly fit real electrophysiological recordings of cortical neurons, where very fine sub-millisecond discretisation steps are important for accurately capturing exact spike timing.
\end{itemize}

\section{Background and related work}
\label{section:background}
\paragraph{Standard ALIF model}
We introduce the recurrent SNN of ALIF neurons with fixed ARP and latency of recurrent transmission, defined by the following set of equations \cite{bellec2018long, yin2021accurateMG}.
\begin{align}
S_i^{(l)}[t]&=f(V^{(l)}_i[t])=\mathbb{1}_{V^{(l)}_i[t] > \theta_i^{(l)}[t]} \quad \text{(Output spike)} \label{eq:f} \\
V^{(l)}_i[t] &= \big(\beta_i^{(l)} V^{(l)}_i[t-1] + (1 -\beta_i^{(l)}) I_i^{(l)}[t] \big) \big(1 - S^{(l)}_i[t-1]\big) \quad \text{(Membrane potential)} \label{eq:disc_lif} \\
\tilde{I}_i^{(l)}[t] &= \Big( b_i^{(l)} + \underbrace{\sum_{j=1}^{N^{(l-1)}} W^{(l)}_{ij} S^{(l-1)}_j[t]}_\text{Feedforward current} + \underbrace{\sum_{j=1}^{N^{(l)}} W^{\text{rec }(l)}_{ij} S^{(l)}_j[t-D]}_\text{Recurrent current} \Big) \quad \text{(Input current)} \label{eq:input_current} \\
I_i^{(l)}[t]&=\begin{cases}
    \tilde{I}_i^{(l)}[t] & \text{if } C_i^{(l)}\geq T_R \\
    0 & \text{otherwise}
\end{cases} \quad \text{(Absolute refractory period)} \label{eq:arp} \\
\begin{split}
	\theta_i^{(l)}[t] &= 1 + d_i^{(l)} a_i^{(l)}[t]\\
	a_i^{(l)}[t] &= p_i^{(l)}a_i^{(l)}[t-1] + S_i^{(l)}[t-1]
	\end{split} \quad \text{(Adaptation)}\label{eq:standard_adapt}
\end{align}

At time $t$, neuron $i$ within layer $l$ (consisting of $N^{(l)}$ neurons) receives input current $I_i^{(l)}[t]$ and outputs a binary value $S_i^{(l)}[t] \in \{ 1, 0\}$ (\textit{i.e.} spike or no spike) if a neuron's membrane potential $V_i^{(l)}[t]$ reaches firing threshold $\theta_i^{(l)}$ (Equation \ref{eq:f}).\footnote{Here notation $\mathbb{1}_\text{\footnotesize condition}$ denotes the indicator function, which is equal to one if the $\text{\footnotesize condition}$ is true and zero otherwise.} The evolution of the membrane potential is described by the normalised discretised LIF equation (Equation \ref{eq:disc_lif}), in which the membrane potential dissipates by a learnable factor $0 \leq \beta_i^{(l)} \leq 1$ at every time step and resets to zero if a spike occurred at the previous time step. The input current is comprised of a constant bias source $b_i^{(l)}$ and from incoming spikes reflecting feedforward $W^{(l)} \in \mathbb{R}^{N^{(l)} \times N^{(l-1)}}$ and recurrent connectivity $W^{\text{rec }(l)} \in \mathbb{R}^{N^{(l)} \times N^{(l)}}$ (Equation \ref{eq:input_current}). 

Here, we assume the recurrent transmission latencies $D$ to be of fixed length and equal in length to the ARP, that is $T_R = D$. With a simple modification however, our methods can work for longer $D$, or different $D$ on each connection, so long as $D \geq T_R$. The ARP is enforced by only allowing input current to enter the neuron if the number of time steps $C_i^{(l)}$ following the last spike equals or exceeds the ARP length (Equation \ref{eq:arp}). Lastly, adaptation is implemented by raising the firing threshold $\theta_i^{(l)}[t]$ following each spike $S_i^{(l)}[t-1]$ (Equation \ref{eq:standard_adapt}), which decays exponentially to baseline $\theta_i^{(l)}=1$ in the absence of any spikes (using learnt decay factor $0 \leq p_i^{(l)} \leq 1$ and adaptation scalar $d_i^{(l)}$). 

The ARP in biological neurons is typically about $\sim 1-2$ms \cite{andersen1978functional, avissar2013refractoriness, rolls1971absolute}. Our method takes advantage of the fact that the monosynaptic connection latency between neurons in local circuits is typically also often around $\sim 1-2$ms \cite{jouhanneau2015vivo}. LIF and ALIF neuronal models are typically run with a DT of about $\sim 0.1$ms in the computational neuroscience literature and $\sim 1$ms in machine learning literature \cite{perezinit}. Furthermore, the firing rate of neurons is often less than the reciprocal of the ARP, suggesting that a higher $T_R$ than the ARP may still provide a reasonable approximation of neural behaviour for some purposes.

\paragraph{SNN training}
The main problem with training SNNs is the non-differentiable nature of their activation function in Equation \ref{eq:f} (whose derivative is undefined at $V^{(l)}_i[t]=\theta_i^{(l)}[t]$ and zero otherwise). This precludes the direct use of the backprop algorithm \citep{rumelhart1986learning}, which has underpinned the successful training of ANNs. A popular solution is to replace the undefined spike derivative with a well-behaved function, referred to as a surrogate gradient \cite{esser2016convolutional, hunsberger2015spiking, zenke2018superspike, lee2016training}, and training the network with backprop-through-time \cite{neftci2019surrogate}. This method also supports the training of neural parameters other than synaptic connectivity, such as membrane time constants \cite{perez2021neural, yin2020effective} and adaptive firing thresholds \cite{bellec2018long, bellec2020solution, yin2020effective}, both of which we include in our work, as they have been shown to improve performance (and increase biological realism). It is worth mentioning that other SNN training methods exist such as mapping trained ANNs to SNNs by transforming ANN neuron activations to SNN neuron firing rates \citep{o2013real, esser2015backpropagation, rueckauer2016theory, rueckauer2017conversion}. These methods are, however, of less interest to computational neuroscientists as they discard all temporal spike precision.

\paragraph{Related work}
Recent work has provided new theoretical insights for increasing the simulation accuracy when employing a larger DT $\geq1$ms \cite{perezinit}. We are not aware of any work that accelerates the simulation and training times on GPUs when employing a smaller DT $\leq1$ms (although see the NEST simulation library for simulating SNNs on CPUs \cite{Gewaltig:NEST, morrison2005advancing, morrison2007exact}). There are, however, different methods for speeding up the backward pass (\textit{i.e.} how gradients are computed within the SNN), which consequentially speeds up training times. Rather than being viewed as competing approaches, these methods could further augment the speed of our solution. In sparse gradient descent, gradients are only passed through neurons whose membrane potential is close to firing threshold, which can significantly accelerate the backward pass when neurons are mostly silent \cite{perez2021sparse}. Inspired by work on training non-spiking networks \cite{williams1989learning, kag2021training}, other methods completely bypass the backward pass and adjust weights online \cite{murray2019local, bellec2020solution, yin2021accurate}. Another approach is to propagate gradient information through spikes \cite{bohte2002error, mostafa2017supervised, comsa2020temporal, kheradpisheh2020temporal, zhang2021rectified, zhou2021spiking, zhou2021temporal}, which - similar to the idea of sparse gradient descent - is fast when neurons are mostly silent. This method, however, enforces neurons to spike at most once and can suffer from training instabilities (although see \cite{goltz2021fast, zhu2022training}).

\section{Theoretical results}
\label{section:theory}
\begin{figure}
    \includegraphics[width=0.9\linewidth]{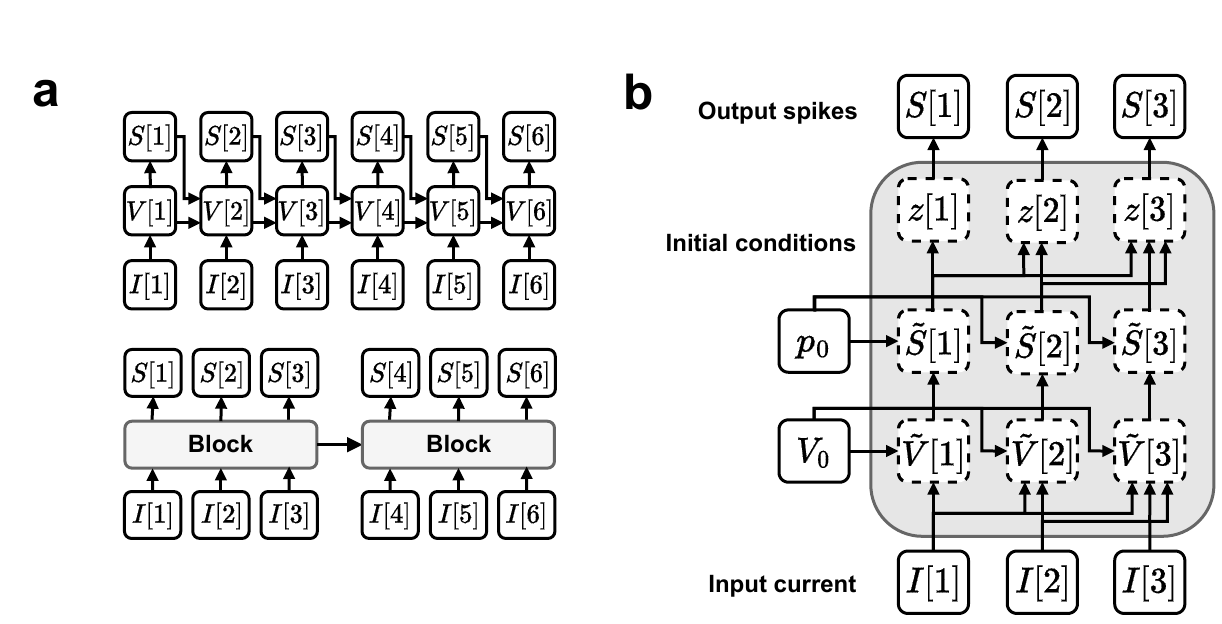}
	\centering
	\caption{\textbf{Solution overview}. \textbf{a.} Our proposed solution: instead of simulating network dynamics one time step after another (top), we sequentially simulate blocks of time equal in length to the neuron ARP (bottom), in which a neuron can spike at most once. \textbf{b.} A schematic of a Block: our proposed solution for emulating ALIF dynamics with a constant sequential complexity $O(1)$ over a short duration in which a neuron spikes at most once.}
 	\label{fig:figure2}
\end{figure}
We outline a novel reformulation of the ALIF model, which theoretically reduces the sequential simulation complexity from $O(T)$ to $O(T/T_R)$ (for simulation length $T$ and ARP length $T_R$). Using the observation that a neuron can spike at most once over simulation length $T_R$ (equal to the ARP; Figure 1c), we propose  simulating network dynamics sequentially in blocks of length $T_R$ - as opposed to simulating every time step individually (Figure \ref{fig:figure2}a) - as we show that these blocks can be simulated with a constant sequential complexity $O(1)$. Our reformulated ALIF model is mathematically the same as the standard ALIF model, but substantially faster to simulate and train. All the proofs for the propositions can be found in the Supplementary material.
	
\subsection{Block: simulating single-spike ALIF dynamics with a constant sequential complexity}
A SNN exhibits a sequential dependence due to the spike reset mechanism, where a neuron's membrane potential $V_i[t]$ can be reset based on its previous output $S_i[t-1]$. Consequently, the simulation of a SNN necessitates a sequential approach. This restriction can, however, be alleviated if we assume a neuron to spike at most once over a particular simulation length (such as the ARP). The following steps - grouped together into a module referred to as a Block (Figure \ref{fig:figure2}b) - compute ALIF dynamics (assuming at most one spike) without any sequential operations.
\begin{align}
\tilde{V}_i[t] &= \big(I_i * \tilde{\beta}_i\big)[t]  \quad \text{(No-reset membrane potential)} \label{eq:block_1} \\
\tilde{S}_i[t] &= f(\tilde{V}_i[t]) \quad \textrm{(Faulty output spikes)} \label{eq:block_2} \\
z_i[t] &= \phi(\tilde{S}_i)[t] \quad \textrm{(Latent timing of spikes)} \label{eq:block_3} \\
S_i[t] &= \mathbb{1}_{\displaystyle z_i[t] = 1} \quad \textrm{(Correct output spikes)} \label{eq:block_4}
\end{align}

\paragraph{1. Calculate membrane potentials without reset} The first step in the Block (Equation \ref{eq:block_1}) converts input current $I_i[t]$ to membrane potentials $\tilde{V}_i[t]$ without spike reset (\textit{i.e.} excluding the reset mechanism in Equation \ref{eq:disc_lif}). This transformation is achieved using a convolution (Proposition \ref{prop:conv_mem}), thus avoiding any sequential operations.
\begin{proposition}
\label{prop:conv_mem}
Membrane potentials without spike reset are computed as a convolution $\tilde{V}_i[t] = \big(I_i * \tilde{\beta}_i\big)[t]$ between input current $I_i[t]$ and kernel $\tilde{\beta}_i[t]=(1-\beta_i)\beta_i^t$ with the initial membrane potential encoded as $I_i[0]=\frac{V_i[0]}{1-\beta_i}$.
\end{proposition}

\paragraph{2. Faulty output spikes} No-reset membrane potentials $\tilde{V}_i[t]$ are mapped to erroneous output spikes $\tilde{S}_i[t]=f(\tilde{V}_i[t])$ (Equation \ref{eq:block_2}) using spike function $f$ (Equation \ref{eq:f}). This output can contain more than one spike, but only the first spike complies with the standard model dynamics, due to the omission of the reset mechanism and ARP constraint. Thus, to ensure that the Block only emits a single spike, all spikes succeeding the first spike occurrence are removed using the following steps.

\paragraph{3. Latent timing of spikes} Erroneous spike output $\tilde{S}_i[t]$ is mapped to a latent representation $z_i[t] = \phi(\tilde{S}_i)[t]$ (Equation \ref{eq:block_3}), encoding the timing of spikes. Function $\phi(\cdot)$ (taking vector $\tilde{S}_i$ as input; Proposition \ref{prop:phi}) is constructed to map all erroneous spikes $\tilde{S}_i[t]$, besides the first spike occurrence, to a value other than one (\textit{i.e.} $z_i[t] \neq 1$ for all $t$ except for the smallest $t$ satisfying $\tilde{S}_i[t]=1$ if such $t$ exists).
\begin{proposition}
\label{prop:phi}
Function $\phi(\tilde{S}_i)[t]=\sum_{k=1}^{t}\tilde{S}_i[k](t-k+1)$ acting on $\tilde{S}_i \in \{0, 1\}^T$ contains at most one element equal to one $\phi(\tilde{S}_i)[t]=1$ for the smallest $t$ satisfying $\tilde{S}_i[t]=1$ (if such $t$ exists).
\end{proposition}

\paragraph{4. Correct output spikes} Lastly, the correct spike output $S_i[t]=\mathbb{1}_{\displaystyle z_i[t] = 1}$ is obtained by setting every value in $z_i[t]$, besides the value one (\textit{i.e.} the first spike), to zero (Equation \ref{eq:block_4}).

\subsection{Blocks: simulating ALIF SNN dynamics with a $O(T/T_R)$ sequential complexity}
The standard ALIF SNN model can be reformulated using a chaining of Blocks, which reduces the sequential simulation complexity (as each Block is simulated in $O(1)$). For a given ARP of length $T_R$, we observed that a neuron spikes at most once over simulation length $T_R$ (Figure \ref{fig:figure1}c). Thus, a simulation length $T$ can be simulated using $N = \frac{T}{T_R}$ Blocks, each of length $T_R$. \footnote{With appropriate zero padding to the simulation length if it is not divisible by $T_R$.} Next, we outline how to simulate ALIF dynamics across Blocks to emulate the dynamics of the standard ALIF SNN. We introduce new notation to index Block $1 \leq n \leq N$ using subscript $n$ (\textit{e.g.} input current to neuron $i$ simulated in Block $n$ is expressed as $I_{i,n}[t]$) with time steps indexed between $[1, T_R]$ (as opposed to $[1, T]$) within a Block.

\paragraph{Input current and the ARP of a Block}
The input current in the standard model (Equation \ref{eq:input_current} and \ref{eq:arp}) is modified for the Block model (Proposition \ref{prop:blocks_input_current}). The feedforward and recurrent current to Block $n+1$ are derived from the presynaptic and postsynaptic spikes from Block $n+1$ and Block $n$ respectively. In addition, the ARP is enforced by applying a mask derived from the latent timing of spikes $z_{i, n}[t]$ (Equation \ref{eq:block_3}).
\begin{proposition}
\label{prop:blocks_input_current}
The input current $I_{i, n+1}[t]$ of neuron $i$ simulated in Block $n+1$ (of length $T_R$) is defined as follows, and enforces an absolute refractory period of length $T_R$ and a monosynaptic transmission latency of $D=T_R$.
\begin{equation}
\nonumber
I_{i, n+1}[t] = \Big( b_{i} + \underbrace{\sum_{j=1}^{N^\text{in}} W_{ij} S_{j, n+1}[t]}_\text{Feedforward current} + \underbrace{\sum_{j=1}^{N^\text{out}} W^{\text{rec}}_{ij} S_{j, n}[t]}_\text{Recurrent current} \Big) \underbrace{\mathbb{1}_{\displaystyle z_{i, n}[t] \geq \max_t S_{i, n}[t]}\vphantom{\sum_{j=1}^{N^{(l)}}}}_\text{ARP mask}
\end{equation}
\end{proposition}

\paragraph{Evolving membrane potentials between Blocks}
Two cases are distinguished to correctly emulate the evolution of the membrane potentials between Blocks (Proposition \ref{prop:blocks_membrane}): 1) if neuron $i$ did not spike in Block $n$ (\textit{i.e.} $\max_t S_{i, n}[t] = 0$), then its initial membrane potential $V_{i, n+1}[0]$ in Block $n+1$ is set to its final membrane potential $V_{i, n}[T_R]$ in Block $n$. Otherwise 2), the initial membrane potential is set to zero to emulate a spike reset (and no state needs to be transferred between Blocks as the neuron is in a refractory state).
\begin{proposition}
\label{prop:blocks_membrane}
The initial membrane potential $V_{i, n+1}[0]$ of neuron $i$ simulated in Block $n+1$ (of length $T_R$) is equal to the last membrane potential in Block $n$ if no spike occurred and zero otherwise.
\begin{equation}
\nonumber
V_{i, n+1}[0]=\begin{cases}
    V_{i, n}[T_R] & \text{if } \max_t S_{i, n}[t] = 0 \\
    0 & \text{otherwise}
\end{cases}
\end{equation} 
\end{proposition}

\paragraph{Evolving adaptive firing thresholds between Blocks}
The adaptive firing threshold $\theta_{i, n+1}[t]$ of neuron $i$ in Block $n+1$ is derived from the initial adaptive parameter $a_{i, n+1}[0]$ (Proposition \ref{prop:blocks_adaptive_thresh}). Two cases are distinguished for deriving this parameter: 1) if the neuron did not spike during the previous Block $n$, this parameter is set to its last value, $a_{i, n}[T_R]$, in the prior Block; otherwise 2), the effect of the spike needs to be taken into account, for which the initial adaptive parameter is expressed as $p_{i}^m (a_s + p_i^{-1})$. Here, $m$ is the number of time steps remaining in Block $n$ following the spike and $a_s$ is the adaptive parameter value at the time of spiking.
\begin{proposition}
\label{prop:blocks_adaptive_thresh}
The adaptive firing threshold $\theta_{i, n+1}[t]$ of neuron $i$ simulated in Block $n+1$ (of length $T_R$) is constructed from the initial adaptive parameter $a_{i, n+1}[0]$, which is equal to its last value in the previous Block if no spike occurred, and otherwise equal to an expression which accounts for the effect of the spike on the adaptive threshold.
\begin{align}
\theta_{i, n+1}[t] &= 1 + d_i p_i^{t} a_{i, n+1}[0] \nonumber\\
a_{i, n+1}[0]&=\begin{cases}
    a_{i, n}[T_R] & \text{if } \max_t {S_{i, n}[t]} = 0 \nonumber\\
    p_{i}^m (a_s + p_i^{-1}) & \text{otherwise}
\end{cases}\\
m = \sum_{k}^{T_R} &\mathbb{1}_{\displaystyle z_{i, n}[k]>1}, \quad a_s = \sum_{k}^{T_R} a_{i, n}[k]  S_{i, n}[k] \nonumber
\end{align}
\end{proposition}

\subsection{Theoretical speedup for simulating ALIF SNNs using Blocks}
\begin{table}[h]
\centering
{\renewcommand{\arraystretch}{1.5}
\begin{tabular}{l c c} 
\hline 
Method & Computational Complexity & Sequential Operations \\ [0.5ex] 
\hline
Standard & $O(N^\text{in} N^\text{out} T)$ & $O(T)$ \\ 
Blocks & $O(N^\text{in} N^\text{out} T_R^2 N)$ & $O(T / T_R)$ \\ [1ex] 
\hline
\end{tabular}}
\caption{Computational and sequential complexity of simulating a layer of $N^\text{out}$ neurons with $N^\text{in}$ input neurons for $T$ time steps using the standard method and our method (with $N$ Blocks each of length $T_R$).}
\label{table:complexity}
\end{table}
Simulating an ALIF SNN using our Blocks, rather than the conventional approach, requires fewer sequential operations (Table \ref{table:complexity}). Although the computational complexity of our Blocks approach is larger than the standard approach, the number of sequential operations is less. If we assume that the sequential steps in both methods are executed in an equal amount of time (as all the non-sequential operations can be run in parallel on a GPU), we obtain a theoretical speedup equal to the length of the ARP $\frac{T}{N}=T_R$.

\section{Experiments}
We evaluated the training speedup of our accelerated ALIF SNN relative to the standard ALIF SNN and explored the capacity of our model to be fitted using different spiking classification datasets and real electrophysiological recordings of cortical neurons. Implementation details can be found in the Supplementary material and the code at \url{https://github.com/webstorms/Blocks}.

\subsection{Training speedup scales with an increasing ARP}
\label{sec:speedup}
\begin{figure}[h]
    \includegraphics[width=1\linewidth]{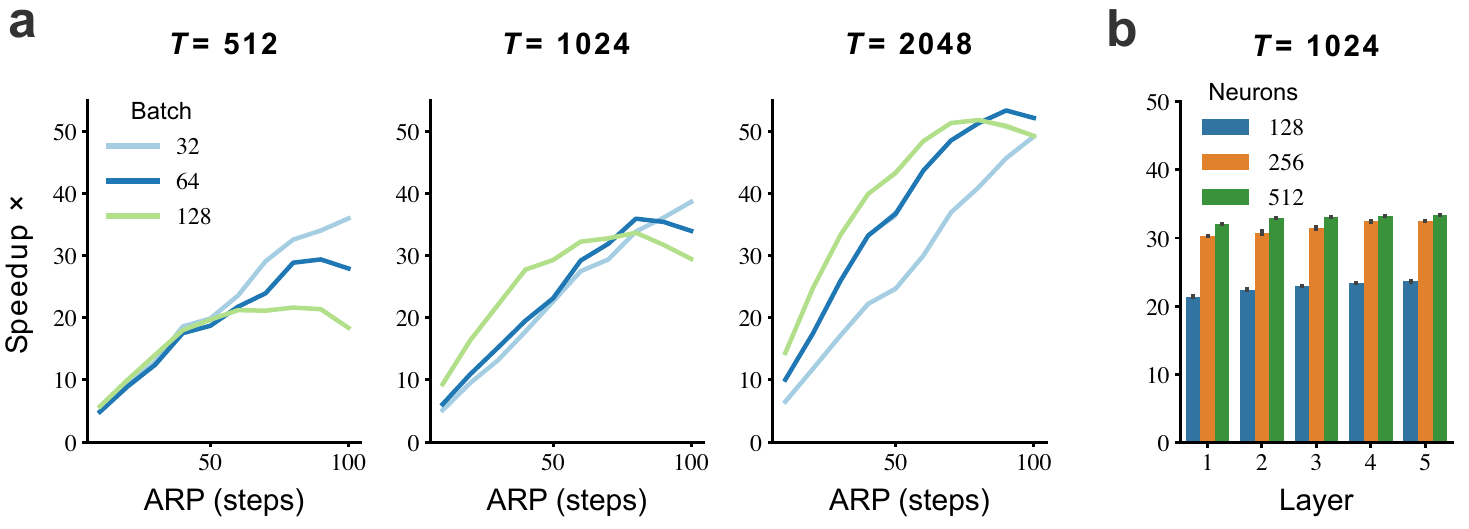}
	\centering
	\caption{\textbf{Training speedup of our model.} \textbf{a.} Training speedup of our accelerated ALIF model compared to the standard ALIF model for different simulation lengths $T$, ARP time steps and batch sizes. \textbf{b.} Training speedup over different number of layers and hidden units (with an ARP of $T_R=40$ time steps, $T=1024$ time steps and batch size $=64$; bars plot the mean and standard error over ten runs). Assuming DT$=0.1$ms, then $10$ time steps $=1$ms.}
 	\label{fig:figure3}
\end{figure}

To determine how much faster our model is, we benchmarked the required training duration (forward and backward pass) of our accelerated ALIF SNN to the standard ALIF SNN for varying ARP and simulation lengths using a synthetic spike dataset, with the task of minimizing the number of final-layer output spikes (see Supplementary material). We found the training speedup of our model to increase for a growing ARP (Figure \ref{fig:figure3}a). This speedup was more pronounced for longer simulation durations ($53\times$ speedup for $T=2048$) than shorter simulation durations ($36\times$ speedup for $T=512$). These speedups were also present when just running the forward pass (\textit{i.e.} inference using the network; see Supplementary material). Furthermore, we found the speedup of our model to be robust over varying numbers of neurons and layers (Figure \ref{fig:figure3}b). Lastly, we also found our method to perform the forward pass  more than an order of magnitude  faster than other publicly available SNN implementations \cite{norse2021, SpikingJelly} (see Supplementary material).
	
\subsection{Accelerated training on spiking classification tasks}
\label{sec:classification}
\begin{figure}[h]
    \includegraphics[width=1\linewidth]{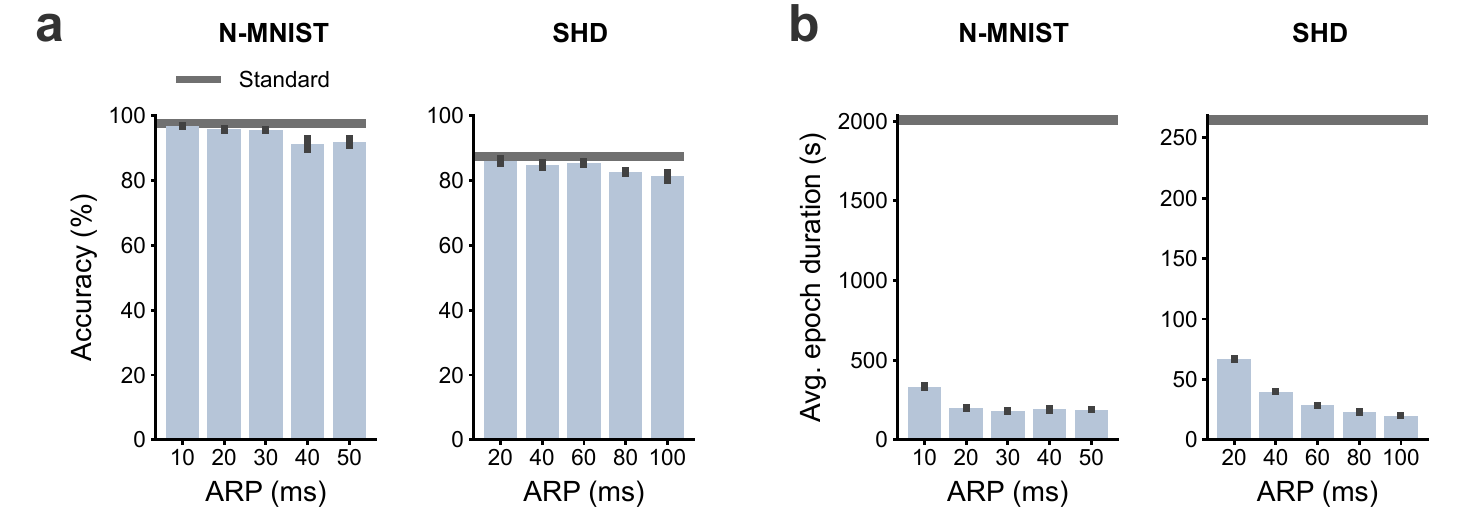}
	\centering
	\caption{\textbf{Performance of our model on spiking datasets.} \textbf{a.} Classification accuracy of our model and the standard ALIF SNN on the N-MNIST and SHD datasets over different (non-biological) ARPs (N-MNIST $1$ms$=$1 time step; SHD $2$ms$=$1 time step). \textbf{b.} Training durations of our model and the standard ALIF SNN on the N-MNIST and SHD datasets. Horizontal gray lines plot the standard model's performance (using no ARP) and bars plot the mean and standard error over three runs.}
 	\label{fig:figure4}
\end{figure}

To establish whether our model can learn using backprop with surrogate gradients, and perform on a par with the standard model, we trained our accelerated ALIF SNN and the standard ALIF SNN on the Neuromophic-MNIST (N-MNIST) (using DT$=1$ms) \cite{orchard2015converting} and Spiking Heidelberg Digits (SHD) (using DT$=2$ms) \cite{cramer2020heidelberg} spiking classification datasets (both commonly used for benchmarking SNN performance \cite{lee2016training, shrestha2018slayer, yin2021accurate, perez2021neural, perez2021sparse}). The task of the N-MNIST dataset is to classify spike representations of handwritten digits, and the task of the SHD dataset is to classify spike representations of spoken digits. In all experiments we employed identical model architectures consisting of two hidden layers (of $256$ neurons each) and an additional integrator readout layer, with predictions taken from the readout neurons with maximal summated membrane potential over time (as commonly done \cite{cramer2020heidelberg, zenke2021remarkable, yin2021accurate, perez2021neural, perez2021sparse}; see Supplementary material). We adopted the multi-Gaussian surrogate-gradient function \cite{yin2021accurate} to overcome the non-differentiable discontinuity of the spike function (although we found that other surrogate-gradient functions also work, see Supplementary material). Furthermore (as suggested by \cite{zenke2021remarkable} for conventional SNNs), we only permitted surrogate gradients to flow through the non-recurrent connectivity (which significantly improved performance; see Supplementary material).

We trained our model with different non-biological ARPs on each dataset and contrasted the accuracy and training duration to the standard model (with no ARP). We found a favourable trade-off between classification accuracy and training time for our model. Compared to the standard ALIF SNN model, our model achieved a very similar, albeit slightly lower, accuracy on the N-MNIST ($96.97\%$ for ARP=$10$ms vs standard $97.71\%$) and SHD dataset ($86.07\%$ for ARP=$20$ms vs standard $87.48\%$), with the accuracy declining only for the largest ARPs tested (Figure \ref{fig:figure4}a). However, our model only required a fraction of the training time, with an average training epoch of $181$s and $20$s for the N-MNIST and SHD datasets, respectively, when using the largest tested ARP, compared to the respective standard model training times of $2034$s and $268$s (Figure \ref{fig:figure4}b).
	
\subsection{Quickly fitting real neural recordings on sub-millisecond timescales}
\label{sec:neural_fit}
\begin{figure}[h]
    \includegraphics[width=0.95\linewidth]{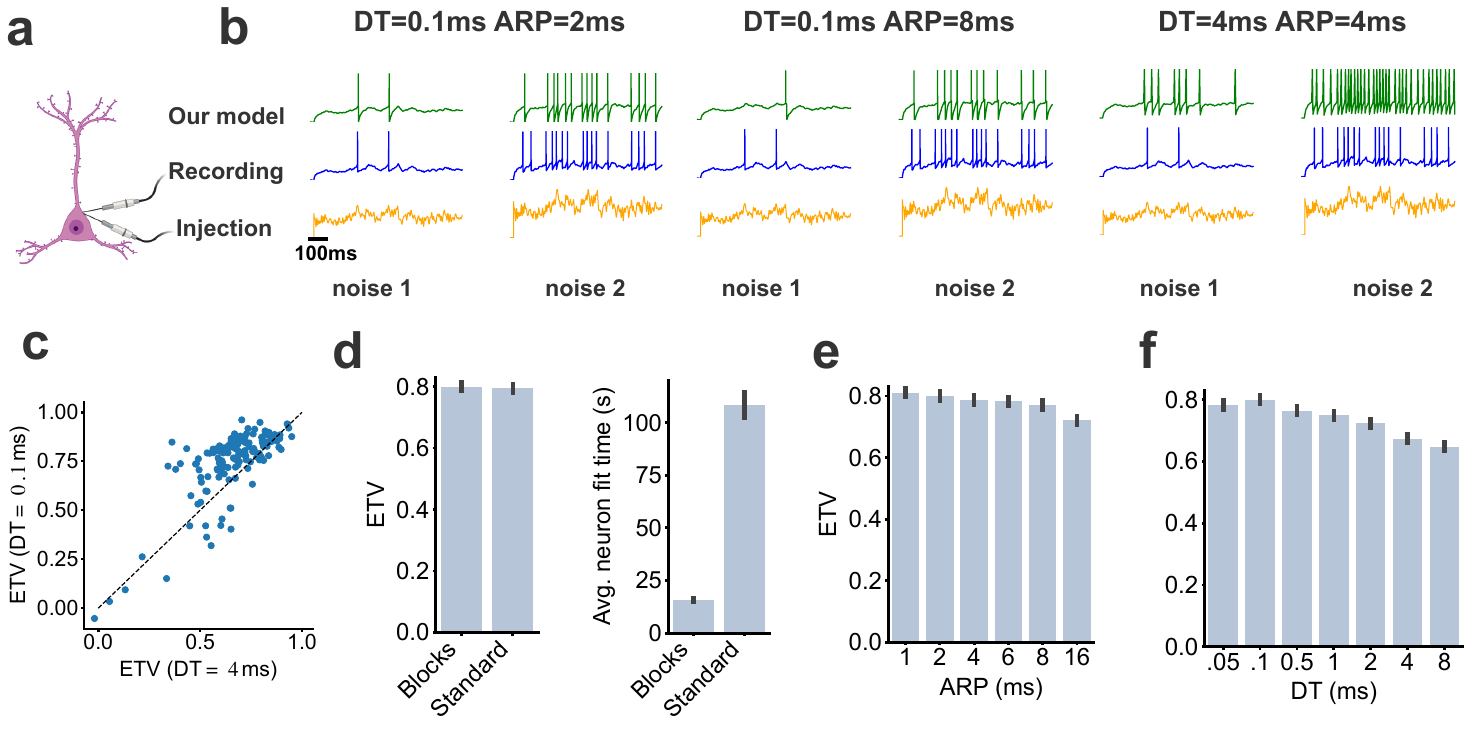}
	\centering
	\caption{\textbf{Fitting cortical electrophysiological recordings.} \textbf{a.} An illustration of a cortical neuron in mouse V1 being recorded whilst stimulated with a noisy current injection. \textbf{b.} For held-out data not used for fitting, an example current injection (bottom) and recorded membrane potential (middle) with corresponding fitted model predictions (top). \textbf{c.} Comparison of neuron fit accuracy of our model for DT$=0.1$ms (y-axis) against DT$=4$ms (x-axis). Explained temporal variance (ETV) measures the goodness-of-fit (one is a perfect fit and zero is a chance-level fit). \textbf{d.} Comparison of the fit accuracy (left) and duration (right) of our model and the standard model (both using DT$=0.1$ms and ARP$=2$ms). \textbf{e.} Our model's fit accuracy for increasing ARP (with DT$=0.1$ms). \textbf{f.} Our model's fit accuracy for increasing DT (with ARP=$\text{max}(2, \text{DT})$ms). \textbf{d.} to \textbf{f.} plots the median and standard error over neurons, except that \textbf{d.} (right) plots the mean fit time.}
 	\label{fig:figure5}
\end{figure}
We explored the ability of our model to fit \textit{in vitro} electrophysiological recordings from $146$ inhibitory and excitatory neurons in mouse primary visual cortex (V1) (provided by the Allen Institute \cite{lein2007genome, hawrylycz2012anatomically}). In these recordings, a variable input current was repeatedly injected at various amplitudes into each real neuron, and the resulting alteration in the membrane potential was recorded (Figure \ref{fig:figure5}a). For each neuron, we used half of the recordings for fitting and the other half for testing, and report all the qualitative and quantitative results on the held-out test dataset. Fitting was achieved by minimising the van Rossum distance between the model and real spike trains \cite{van2001novel}. To quantitatively compare the fits of our model using different DTs and ARPs, we employed the explained temporal variance (ETV) measure (used on the Allen Institute website; see Supplementary material). This metric quantifies how well the temporal dynamics of the neurons are captured by the model and takes into account trial-to-trial variability due to intrinsic neural noise. It has a value of one for a perfect fit and a value of zero if the model predicts at chance.

We found that our accelerated ALIF model captured the spike timing of the neurons. Pertinent to the speed-accuracy trade-off, this fit strongly depended on the chosen DT, and less so on the chosen ARP. Qualitatively, using a DT$=0.1$ms and an ARP$=2$ms, we found our model captured the spike timings of the neurons for current injections of varying amplitude. This still seemed to be the case when we used a larger ARP$=8$ms, but the model was worse at capturing spike timings when we used a larger DT$=4$ms (with ARP$=4$ms; Figure \ref{fig:figure5}b; see Supplementary material for zoomed-in neural traces). 

Quantitatively comparing the neuron fits one by one, we found that nearly all of the neuron fits were better when using a DT$=0.1$ms (ETV$=0.80$) than a DT$=4$ms (ETV$=0.66$; Figure \ref{fig:figure5}c; using an ARP$=4$ms). We examined how accurate and fast our fits were compared to the standard model using an ARP$=2$ms and a DT$=0.1$ms. Both models achieved a similar ETV of $\sim 0.8$, yet our model only required $15.5$s on average per neuron fit time compared to the $108.4$s of the standard model (Figure \ref{fig:figure5}d). We further investigated whether a larger ARP could still reasonably fit the data for DT$=0.1$ms (to benefit from a faster fit). Consistent with our qualitative observations, we found a less marked reduction in fit accuracy when using larger ARPs (Figure \ref{fig:figure5}e) compared to the drop in performance when using larger DTs (Figure \ref{fig:figure5}f; using an ARP of $\text{max}(2, \text{DT})$ms).

\section{Discussion}
\vspace{-2truemm}
ALIF neurons are a popular model for studying the brain \textit{in silico}. Training these neurons is, however, slow due to their sequential nature \cite{neftci2019surrogate, eshraghian2021training}. We overcome this by algorithmically reinterpreting the ALIF model. We found that our method permits faster simulation of SNNs, which will likely play an important role in future research into neural dynamics through simulation \cite{vogels2011inhibitory, deneve2016efficient}. We also confirmed the validity of this approach for modelling neural data. Firstly - of interest to computational neuroscientists - we fitted a multilayered network of neurons using two common spike classification datasets. We found that our model achieved a similar classification accuracy to that of the standard model, even if we increased the ARP to large non-physiological values, which drastically reduced the training duration required for both datasets. Secondly - of interest to computational and experimental neuroscientists - we explored the applicability of our method to fit real electrophysiological recordings of neurons in mouse V1 using sub-millisecond DTs. 

We found that our method accurately captured the spike timing of real neurons, fitting their activity patterns in a fraction of the time required by the standard model (although we note that other, more recent spiking-models, might improve fit accuracies \cite{pozzorini2015automated}). This is particularly important as datasets become larger with advances in recording techniques, requiring faster fitting solutions to characterise computational properties of neurons. As an example of the potential insights provided by our model, we found the fitted V1 neurons to have a heterogenous membrane time constant distribution, suggesting that V1 processes visual information on multiple timescales \cite{perez2021neural} (see Supplementary material).

Our work will likely also be of interest to neuromorphic engineers and researchers, developing energy-efficient hardware to emulate SNNs \cite{wunderlich2019demonstrating}. These systems - like real neurons - run in continuous time and thus require training on extremely fine DTs off-chip \cite{stuijt2021mubrain, he202128}. Our method could help to accelerate off-chip training times. Furthermore, the ARP hyperparameter in our model can limit high spike rates and thus reduce energy consumption on neuromorphic systems, as this consumption scales approximately proportionally with the number of emitted spikes \citep{panda2020toward}.

A limitation of our method is that the training speedup scales sublinearly - as opposed to linearly - with an increasing ARP simulation length (see Section \ref{section:theory} and \ref{sec:speedup}). This is likely due to GPU overheads and employed cudnn routines \cite{chetlur2014cudnn}, which further improvements to our code implementation could overcome. An additional limitation is the requirement to define the ARP hyperparameter, whose value influences the training speed and test accuracy in our method and relates to biological realism. However, we found a beneficial trade-off - by using large non-biological ARPs on the artificial spiking classification dataset and small physiological ARPs for the neural fits (although we found larger values to also perform reasonably well) the model achieved comparable accuracy to the standard ALIF model in both cases, while also having greatly increased speed of training and simulation. In particular, the capacity of our model to accurately and quickly fit the data from real neurons is crucial in terms of the biological applicability of this approach. 

\begin{ack}
We thank Rob Pratt and anonymous reviewers for helpful discussions; and Lorenzo Mazzaschi for feedback on the manuscript. Luke Taylor was supported by the Clarendon Fund. Andrew King and Nicol Harper were supported by the Wellcome Trust (WT108369/Z/2015/Z). Figure \ref{fig:figure5} was created with BioRender.com.
\end{ack}

\bibliography{bibfile}

\begin{thebibliography}{77}
\providecommand{\natexlab}[1]{#1}
\providecommand{\url}[1]{\texttt{#1}}
\expandafter\ifx\csname urlstyle\endcsname\relax
  \providecommand{\doi}[1]{doi: #1}\else
  \providecommand{\doi}{doi: \begingroup \urlstyle{rm}\Url}\fi

\bibitem[Harper et~al.(2016)Harper, Schoppe, Willmore, Cui, Schnupp, and
  King]{harper2016network}
Nicol~S Harper, Oliver Schoppe, Ben~DB Willmore, Zhanfeng Cui, Jan~WH Schnupp,
  and Andrew~J King.
\newblock Network receptive field modeling reveals extensive integration and
  multi-feature selectivity in auditory cortical neurons.
\newblock \emph{PLoS Computational Biology}, 12\penalty0 (11):\penalty0
  e1005113, 2016.

\bibitem[Singer et~al.(2023)Singer, Taylor, Willmore, King, and
  Harper]{singer2023hierarchical}
Yosef Singer, Luke Taylor, Ben~DB Willmore, Andrew~J King, and Nicol~S Harper.
\newblock Hierarchical temporal prediction captures motion processing along the
  visual pathway.
\newblock \emph{eLife}, 12:\penalty0 e52599, 2023.

\bibitem[Cadena et~al.(2019)Cadena, Denfield, Walker, Gatys, Tolias, Bethge,
  and Ecker]{cadena2019deep}
Santiago~A Cadena, George~H Denfield, Edgar~Y Walker, Leon~A Gatys, Andreas~S
  Tolias, Matthias Bethge, and Alexander~S Ecker.
\newblock Deep convolutional models improve predictions of macaque v1 responses
  to natural images.
\newblock \emph{PLoS Computational Biology}, 15\penalty0 (4):\penalty0
  e1006897, 2019.

\bibitem[Francl and McDermott(2022)]{francl2022deep}
Andrew Francl and Josh~H McDermott.
\newblock Deep neural network models of sound localization reveal how
  perception is adapted to real-world environments.
\newblock \emph{Nature Human Behaviour}, 6\penalty0 (1):\penalty0 111--133,
  2022.

\bibitem[Yamins and DiCarlo(2016)]{yamins2016using}
Daniel~LK Yamins and James~J DiCarlo.
\newblock Using goal-driven deep learning models to understand sensory cortex.
\newblock \emph{Nature Neuroscience}, 19\penalty0 (3):\penalty0 356--365, 2016.

\bibitem[Bakhtiari et~al.(2021)Bakhtiari, Mineault, Lillicrap, Pack, and
  Richards]{bakhtiari2021functional}
Shahab Bakhtiari, Patrick Mineault, Timothy Lillicrap, Christopher Pack, and
  Blake Richards.
\newblock The functional specialization of visual cortex emerges from training
  parallel pathways with self-supervised predictive learning.
\newblock \emph{Advances in Neural Information Processing Systems},
  34:\penalty0 25164--25178, 2021.

\bibitem[Mineault et~al.(2021)Mineault, Bakhtiari, Richards, and
  Pack]{mineault2021your}
Patrick Mineault, Shahab Bakhtiari, Blake Richards, and Christopher Pack.
\newblock Your head is there to move you around: Goal-driven models of the
  primate dorsal pathway.
\newblock \emph{Advances in Neural Information Processing Systems},
  34:\penalty0 28757--28771, 2021.

\bibitem[Ocko et~al.(2018)Ocko, Lindsey, Ganguli, and Deny]{ocko2018emergence}
Samuel Ocko, Jack Lindsey, Surya Ganguli, and Stephane Deny.
\newblock The emergence of multiple retinal cell types through efficient coding
  of natural movies.
\newblock \emph{Advances in Neural Information Processing Systems}, 31, 2018.

\bibitem[Conwell et~al.(2021)Conwell, Mayo, Barbu, Buice, Alvarez, and
  Katz]{conwell2021neural}
Colin Conwell, David Mayo, Andrei Barbu, Michael Buice, George Alvarez, and
  Boris Katz.
\newblock Neural regression, representational similarity, model zoology \&
  neural taskonomy at scale in rodent visual cortex.
\newblock \emph{Advances in Neural Information Processing Systems},
  34:\penalty0 5590--5607, 2021.

\bibitem[Richards et~al.(2019)Richards, Lillicrap, Beaudoin, Bengio, Bogacz,
  Christensen, Clopath, Costa, de~Berker, Ganguli, et~al.]{richards2019deep}
Blake~A Richards, Timothy~P Lillicrap, Philippe Beaudoin, Yoshua Bengio, Rafal
  Bogacz, Amelia Christensen, Claudia Clopath, Rui~Ponte Costa, Archy
  de~Berker, Surya Ganguli, et~al.
\newblock A deep learning framework for neuroscience.
\newblock \emph{Nature Neuroscience}, 22\penalty0 (11):\penalty0 1761--1770,
  2019.

\bibitem[Jolivet et~al.(2008)Jolivet, Sch{\"u}rmann, Berger, Naud, Gerstner,
  and Roth]{jolivet2008quantitative}
Renaud Jolivet, Felix Sch{\"u}rmann, Thomas~K Berger, Richard Naud, Wulfram
  Gerstner, and Arnd Roth.
\newblock The quantitative single-neuron modeling competition.
\newblock \emph{Biological Cybernetics}, 99:\penalty0 417--426, 2008.

\bibitem[Kobayashi et~al.(2009)Kobayashi, Tsubo, and
  Shinomoto]{kobayashi2009made}
Ryota Kobayashi, Yasuhiro Tsubo, and Shigeru Shinomoto.
\newblock Made-to-order spiking neuron model equipped with a multi-timescale
  adaptive threshold.
\newblock \emph{Frontiers in Computational Neuroscience}, page~9, 2009.

\bibitem[Rossant et~al.(2011)Rossant, Goodman, Fontaine, Platkiewicz,
  Magnusson, and Brette]{rossant2011fitting}
Cyrille Rossant, Dan~FM Goodman, Bertrand Fontaine, Jonathan Platkiewicz,
  Anna~K Magnusson, and Romain Brette.
\newblock Fitting neuron models to spike trains.
\newblock \emph{Frontiers in Neuroscience}, 5:\penalty0 9, 2011.

\bibitem[Mensi et~al.(2012)Mensi, Naud, Pozzorini, Avermann, Petersen, and
  Gerstner]{mensi2012parameter}
Skander Mensi, Richard Naud, Christian Pozzorini, Michael Avermann, Carl~CH
  Petersen, and Wulfram Gerstner.
\newblock Parameter extraction and classification of three cortical neuron
  types reveals two distinct adaptation mechanisms.
\newblock \emph{Journal of Neurophysiology}, 107\penalty0 (6):\penalty0
  1756--1775, 2012.

\bibitem[Pozzorini et~al.(2013)Pozzorini, Naud, Mensi, and
  Gerstner]{pozzorini2013temporal}
Christian Pozzorini, Richard Naud, Skander Mensi, and Wulfram Gerstner.
\newblock Temporal whitening by power-law adaptation in neocortical neurons.
\newblock \emph{Nature Neuroscience}, 16\penalty0 (7):\penalty0 942--948, 2013.

\bibitem[Den{\`e}ve and Machens(2016)]{deneve2016efficient}
Sophie Den{\`e}ve and Christian~K Machens.
\newblock Efficient codes and balanced networks.
\newblock \emph{Nature Neuroscience}, 19\penalty0 (3):\penalty0 375--382, 2016.

\bibitem[Vogels et~al.(2011)Vogels, Sprekeler, Zenke, Clopath, and
  Gerstner]{vogels2011inhibitory}
Tim~P Vogels, Henning Sprekeler, Friedemann Zenke, Claudia Clopath, and Wulfram
  Gerstner.
\newblock Inhibitory plasticity balances excitation and inhibition in sensory
  pathways and memory networks.
\newblock \emph{Science}, 334\penalty0 (6062):\penalty0 1569--1573, 2011.

\bibitem[Confavreux et~al.(2020)Confavreux, Zenke, Agnes, Lillicrap, and
  Vogels]{confavreux2020meta}
Basile Confavreux, Friedemann Zenke, Everton Agnes, Timothy Lillicrap, and Tim
  Vogels.
\newblock A meta-learning approach to (re) discover plasticity rules that carve
  a desired function into a neural network.
\newblock \emph{Advances in Neural Information Processing Systems},
  33:\penalty0 16398--16408, 2020.

\bibitem[Braun and Vogels(2021)]{braun2021online}
Lukas Braun and Tim Vogels.
\newblock Online learning of neural computations from sparse temporal feedback.
\newblock \emph{Advances in Neural Information Processing Systems},
  34:\penalty0 16437--16450, 2021.

\bibitem[Wunderlich et~al.(2019)Wunderlich, Kungl, M{\"u}ller, Hartel,
  Stradmann, Aamir, Gr{\"u}bl, Heimbrecht, Schreiber, St{\"o}ckel,
  et~al.]{wunderlich2019demonstrating}
Timo Wunderlich, Akos~F Kungl, Eric M{\"u}ller, Andreas Hartel, Yannik
  Stradmann, Syed~Ahmed Aamir, Andreas Gr{\"u}bl, Arthur Heimbrecht, Korbinian
  Schreiber, David St{\"o}ckel, et~al.
\newblock Demonstrating advantages of neuromorphic computation: a pilot study.
\newblock \emph{Frontiers in Neuroscience}, 13:\penalty0 260, 2019.

\bibitem[Lapicque(1907)]{lapicque1907recherches}
Louis Lapicque.
\newblock Recherches quantitatives sur l'excitation electrique des nerfs
  traitee comme une polarization.
\newblock \emph{Journal de physiologie et de pathologie g{\'e}n{\'e}rale},
  9:\penalty0 620--635, 1907.

\bibitem[Gerstner et~al.(2014)Gerstner, Kistler, Naud, and
  Paninski]{gerstner2014neuronal}
Wulfram Gerstner, Werner~M Kistler, Richard Naud, and Liam Paninski.
\newblock \emph{Neuronal dynamics: From single neurons to networks and models
  of cognition}.
\newblock Cambridge University Press, 2014.

\bibitem[Kandel et~al.(2000)Kandel, Schwartz, Jessell, Siegelbaum, Hudspeth,
  Mack, et~al.]{kandel2000principles}
Eric~R Kandel, James~H Schwartz, Thomas~M Jessell, Steven Siegelbaum, A~James
  Hudspeth, Sarah Mack, et~al.
\newblock \emph{Principles of neural science, Fourth edition}.
\newblock Elsevier, 2000.

\bibitem[Levakova et~al.(2019)Levakova, Kostal, Monsemp{\`e}s, Lucas, and
  Kobayashi]{levakova2019adaptive}
Marie Levakova, Lubomir Kostal, Christelle Monsemp{\`e}s, Philippe Lucas, and
  Ryota Kobayashi.
\newblock Adaptive integrate-and-fire model reproduces the dynamics of
  olfactory receptor neuron responses in a moth.
\newblock \emph{Journal of the Royal Society Interface}, 16\penalty0
  (157):\penalty0 20190246, 2019.

\bibitem[Zeng et~al.(2021)Zeng, Stewart, Ibne~Ferdous, Berdichevsky, and
  Guo]{zeng2021temporal}
Yuan Zeng, Terrence~C Stewart, Zubayer Ibne~Ferdous, Yevgeny Berdichevsky, and
  Xiaochen Guo.
\newblock Temporal learning with biologically fitted snn models.
\newblock In \emph{International Conference on Neuromorphic Systems 2021},
  pages 1--8, 2021.

\bibitem[Bellec et~al.(2018)Bellec, Salaj, Subramoney, Legenstein, and
  Maass]{bellec2018long}
Guillaume Bellec, Darjan Salaj, Anand Subramoney, Robert Legenstein, and
  Wolfgang Maass.
\newblock Long short-term memory and learning-to-learn in networks of spiking
  neurons.
\newblock \emph{Advances in neural information processing systems}, 31, 2018.

\bibitem[Bellec et~al.(2020)Bellec, Scherr, Subramoney, Hajek, Salaj,
  Legenstein, and Maass]{bellec2020solution}
Guillaume Bellec, Franz Scherr, Anand Subramoney, Elias Hajek, Darjan Salaj,
  Robert Legenstein, and Wolfgang Maass.
\newblock A solution to the learning dilemma for recurrent networks of spiking
  neurons.
\newblock \emph{Nature Communications}, 11\penalty0 (1):\penalty0 1--15, 2020.

\bibitem[Yin et~al.(2020)Yin, Corradi, and Boht{\'e}]{yin2020effective}
Bojian Yin, Federico Corradi, and Sander~M Boht{\'e}.
\newblock Effective and efficient computation with multiple-timescale spiking
  recurrent neural networks.
\newblock In \emph{International Conference on Neuromorphic Systems 2020},
  pages 1--8, 2020.

\bibitem[Yin et~al.(2023)Yin, Corradi, and Boht{\'e}]{yin2021accurate}
Bojian Yin, Federico Corradi, and Sander~M Boht{\'e}.
\newblock Accurate online training of dynamical spiking neural networks through
  forward propagation through time.
\newblock \emph{Nature Machine Intelligence}, pages 1--10, 2023.

\bibitem[Perez-Nieves and Goodman(2023)]{perezinit}
Nicolas Perez-Nieves and Dan~FM Goodman.
\newblock Spiking network initialisation and firing rate collapse.
\newblock \emph{arXiv:2305.08879}, 2023.

\bibitem[Yin et~al.(2021)Yin, Corradi, and Boht{\'e}]{yin2021accurateMG}
Bojian Yin, Federico Corradi, and Sander~M Boht{\'e}.
\newblock Accurate and efficient time-domain classification with adaptive
  spiking recurrent neural networks.
\newblock \emph{Nature Machine Intelligence}, 3\penalty0 (10):\penalty0
  905--913, 2021.

\bibitem[Andersen et~al.(1978)Andersen, Silfvenius, Sundberg, Sveen, Wigstro,
  et~al.]{andersen1978functional}
P~Andersen, H~Silfvenius, SH~Sundberg, On~Sveen, H~Wigstro, et~al.
\newblock Functional characteristics of unmyelinated fibres in the hippocampal
  cortex.
\newblock \emph{Brain research}, 144\penalty0 (1):\penalty0 11--18, 1978.

\bibitem[Avissar et~al.(2013)Avissar, Wittig, Saunders, and
  Parsons]{avissar2013refractoriness}
Michael Avissar, John~H Wittig, James~C Saunders, and Thomas~D Parsons.
\newblock Refractoriness enhances temporal coding by auditory nerve fibers.
\newblock \emph{Journal of Neuroscience}, 33\penalty0 (18):\penalty0
  7681--7690, 2013.

\bibitem[Rolls(1971)]{rolls1971absolute}
Edmund~T Rolls.
\newblock Absolute refractory period of neurons involved in mfb
  self-stimulation.
\newblock \emph{Physiology \& Behavior}, 7\penalty0 (3):\penalty0 311--315,
  1971.

\bibitem[Jouhanneau et~al.(2015)Jouhanneau, Kremkow, Dorrn, and
  Poulet]{jouhanneau2015vivo}
Jean-S{\'e}bastien Jouhanneau, Jens Kremkow, Anja~L Dorrn, and James~FA Poulet.
\newblock In vivo monosynaptic excitatory transmission between layer 2 cortical
  pyramidal neurons.
\newblock \emph{Cell reports}, 13\penalty0 (10):\penalty0 2098--2106, 2015.

\bibitem[Rumelhart et~al.(1986)Rumelhart, Hinton, and
  Williams]{rumelhart1986learning}
David~E Rumelhart, Geoffrey~E Hinton, and Ronald~J Williams.
\newblock Learning representations by back-propagating errors.
\newblock \emph{Nature}, 323\penalty0 (6088):\penalty0 533--536, 1986.

\bibitem[Esser et~al.(2016)Esser, Merolla, Arthur, Cassidy, Appuswamy,
  Andreopoulos, Berg, McKinstry, Melano, Barch, et~al.]{esser2016convolutional}
Steven~K Esser, Paul~A Merolla, John~V Arthur, Andrew~S Cassidy, Rathinakumar
  Appuswamy, Alexander Andreopoulos, David~J Berg, Jeffrey~L McKinstry, Timothy
  Melano, Davis~R Barch, et~al.
\newblock Convolutional networks for fast, energy-efficient neuromorphic
  computing.
\newblock \emph{Proceedings of the National Academy of Sciences}, 113\penalty0
  (41):\penalty0 11441--11446, 2016.

\bibitem[Hunsberger and Eliasmith(2015)]{hunsberger2015spiking}
Eric Hunsberger and Chris Eliasmith.
\newblock Spiking deep networks with lif neurons.
\newblock \emph{arXiv:1510.08829}, 2015.

\bibitem[Zenke and Ganguli(2018)]{zenke2018superspike}
Friedemann Zenke and Surya Ganguli.
\newblock Superspike: Supervised learning in multilayer spiking neural
  networks.
\newblock \emph{Neural Computation}, 30\penalty0 (6):\penalty0 1514--1541,
  2018.

\bibitem[Lee et~al.(2016)Lee, Delbruck, and Pfeiffer]{lee2016training}
Jun~Haeng Lee, Tobi Delbruck, and Michael Pfeiffer.
\newblock Training deep spiking neural networks using backpropagation.
\newblock \emph{Frontiers in Neuroscience}, 10:\penalty0 508, 2016.

\bibitem[Neftci et~al.(2019)Neftci, Mostafa, and Zenke]{neftci2019surrogate}
Emre~O Neftci, Hesham Mostafa, and Friedemann Zenke.
\newblock Surrogate gradient learning in spiking neural networks: Bringing the
  power of gradient-based optimization to spiking neural networks.
\newblock \emph{IEEE Signal Processing Magazine}, 36\penalty0 (6):\penalty0
  51--63, 2019.

\bibitem[Perez-Nieves et~al.(2021)Perez-Nieves, Leung, Dragotti, and
  Goodman]{perez2021neural}
Nicolas Perez-Nieves, Vincent~CH Leung, Pier~Luigi Dragotti, and Dan~FM
  Goodman.
\newblock Neural heterogeneity promotes robust learning.
\newblock \emph{Nature communications}, 12\penalty0 (1):\penalty0 1--9, 2021.

\bibitem[O'Connor et~al.(2013)O'Connor, Neil, Liu, Delbruck, and
  Pfeiffer]{o2013real}
Peter O'Connor, Daniel Neil, Shih-Chii Liu, Tobi Delbruck, and Michael
  Pfeiffer.
\newblock Real-time classification and sensor fusion with a spiking deep belief
  network.
\newblock \emph{Frontiers in Neuroscience}, 7:\penalty0 178, 2013.

\bibitem[Esser et~al.(2015)Esser, Appuswamy, Merolla, Arthur, and
  Modha]{esser2015backpropagation}
Steve~K Esser, Rathinakumar Appuswamy, Paul Merolla, John~V Arthur, and
  Dharmendra~S Modha.
\newblock Backpropagation for energy-efficient neuromorphic computing.
\newblock \emph{Advances in neural information processing systems}, 28, 2015.

\bibitem[Rueckauer et~al.(2016)Rueckauer, Lungu, Hu, and
  Pfeiffer]{rueckauer2016theory}
Bodo Rueckauer, Iulia-Alexandra Lungu, Yuhuang Hu, and Michael Pfeiffer.
\newblock Theory and tools for the conversion of analog to spiking
  convolutional neural networks.
\newblock \emph{arXiv:1612.04052}, 2016.

\bibitem[Rueckauer et~al.(2017)Rueckauer, Lungu, Hu, Pfeiffer, and
  Liu]{rueckauer2017conversion}
Bodo Rueckauer, Iulia-Alexandra Lungu, Yuhuang Hu, Michael Pfeiffer, and
  Shih-Chii Liu.
\newblock Conversion of continuous-valued deep networks to efficient
  event-driven networks for image classification.
\newblock \emph{Frontiers in Neuroscience}, 11:\penalty0 682, 2017.

\bibitem[Gewaltig and Diesmann(2007)]{Gewaltig:NEST}
Marc-Oliver Gewaltig and Markus Diesmann.
\newblock Nest (neural simulation tool).
\newblock \emph{Scholarpedia}, 2\penalty0 (4):\penalty0 1430, 2007.

\bibitem[Morrison et~al.(2005)Morrison, Mehring, Geisel, Aertsen, and
  Diesmann]{morrison2005advancing}
Abigail Morrison, Carsten Mehring, Theo Geisel, AD~Aertsen, and Markus
  Diesmann.
\newblock Advancing the boundaries of high-connectivity network simulation with
  distributed computing.
\newblock \emph{Neural computation}, 17\penalty0 (8):\penalty0 1776--1801,
  2005.

\bibitem[Morrison et~al.(2007)Morrison, Straube, Plesser, and
  Diesmann]{morrison2007exact}
Abigail Morrison, Sirko Straube, Hans~Ekkehard Plesser, and Markus Diesmann.
\newblock Exact subthreshold integration with continuous spike times in
  discrete-time neural network simulations.
\newblock \emph{Neural Computation}, 19\penalty0 (1):\penalty0 47--79, 2007.

\bibitem[Perez-Nieves and Goodman(2021)]{perez2021sparse}
Nicolas Perez-Nieves and Dan Goodman.
\newblock Sparse spiking gradient descent.
\newblock \emph{Advances in Neural Information Processing Systems},
  34:\penalty0 11795--11808, 2021.

\bibitem[Williams and Zipser(1989)]{williams1989learning}
Ronald~J Williams and David Zipser.
\newblock A learning algorithm for continually running fully recurrent neural
  networks.
\newblock \emph{Neural Computation}, 1\penalty0 (2):\penalty0 270--280, 1989.

\bibitem[Kag and Saligrama(2021)]{kag2021training}
Anil Kag and Venkatesh Saligrama.
\newblock Training recurrent neural networks via forward propagation through
  time.
\newblock In \emph{International Conference on Machine Learning}, pages
  5189--5200. PMLR, 2021.

\bibitem[Murray(2019)]{murray2019local}
James~M Murray.
\newblock Local online learning in recurrent networks with random feedback.
\newblock \emph{Elife}, 8:\penalty0 e43299, 2019.

\bibitem[Bohte et~al.(2002)Bohte, Kok, and La~Poutre]{bohte2002error}
Sander~M Bohte, Joost~N Kok, and Han La~Poutre.
\newblock Error-backpropagation in temporally encoded networks of spiking
  neurons.
\newblock \emph{Neurocomputing}, 48\penalty0 (1-4):\penalty0 17--37, 2002.

\bibitem[Mostafa(2017)]{mostafa2017supervised}
Hesham Mostafa.
\newblock Supervised learning based on temporal coding in spiking neural
  networks.
\newblock \emph{IEEE transactions on neural networks and learning systems},
  29\penalty0 (7):\penalty0 3227--3235, 2017.

\bibitem[Comsa et~al.(2020)Comsa, Potempa, Versari, Fischbacher, Gesmundo, and
  Alakuijala]{comsa2020temporal}
Iulia~M Comsa, Krzysztof Potempa, Luca Versari, Thomas Fischbacher, Andrea
  Gesmundo, and Jyrki Alakuijala.
\newblock Temporal coding in spiking neural networks with alpha synaptic
  function.
\newblock In \emph{ICASSP 2020-2020 IEEE International Conference on Acoustics,
  Speech and Signal Processing (ICASSP)}, pages 8529--8533. IEEE, 2020.

\bibitem[Kheradpisheh and Masquelier(2020)]{kheradpisheh2020temporal}
Saeed~Reza Kheradpisheh and Timoth{\'e}e Masquelier.
\newblock Temporal backpropagation for spiking neural networks with one spike
  per neuron.
\newblock \emph{International Journal of Neural Systems}, 30\penalty0
  (06):\penalty0 2050027, 2020.

\bibitem[Zhang et~al.(2021)Zhang, Wang, Wu, Belatreche, Amornpaisannon, Zhang,
  Miriyala, Qu, Chua, Carlson, et~al.]{zhang2021rectified}
Malu Zhang, Jiadong Wang, Jibin Wu, Ammar Belatreche, Burin Amornpaisannon,
  Zhixuan Zhang, Venkata Pavan~Kumar Miriyala, Hong Qu, Yansong Chua, Trevor~E
  Carlson, et~al.
\newblock Rectified linear postsynaptic potential function for backpropagation
  in deep spiking neural networks.
\newblock \emph{IEEE Transactions on Neural Networks and Learning Systems},
  33\penalty0 (5):\penalty0 1947--1958, 2021.

\bibitem[Zhou and Li(2021)]{zhou2021spiking}
Shibo Zhou and Xiaohua Li.
\newblock Spiking neural networks with single-spike temporal-coded neurons for
  network intrusion detection.
\newblock In \emph{2020 25th International Conference on Pattern Recognition
  (ICPR)}, pages 8148--8155. IEEE, 2021.

\bibitem[Zhou et~al.(2021)Zhou, Li, Chen, Chandrasekaran, and
  Sanyal]{zhou2021temporal}
Shibo Zhou, Xiaohua Li, Ying Chen, Sanjeev~T Chandrasekaran, and Arindam
  Sanyal.
\newblock Temporal-coded deep spiking neural network with easy training and
  robust performance.
\newblock In \emph{Proceedings of the AAAI Conference on Artificial
  Intelligence}, volume~35, pages 11143--11151, 2021.

\bibitem[G{\"o}ltz et~al.(2021)G{\"o}ltz, Kriener, Baumbach, Billaudelle,
  Breitwieser, Cramer, Dold, Kungl, Senn, Schemmel, et~al.]{goltz2021fast}
Julian G{\"o}ltz, Laura Kriener, Andreas Baumbach, Sebastian Billaudelle,
  Oliver Breitwieser, Benjamin Cramer, Dominik Dold, Akos~Ferenc Kungl, Walter
  Senn, Johannes Schemmel, et~al.
\newblock Fast and energy-efficient neuromorphic deep learning with first-spike
  times.
\newblock \emph{Nature Machine Intelligence}, 3\penalty0 (9):\penalty0
  823--835, 2021.

\bibitem[Zhu et~al.(2022)Zhu, Yu, Fang, Xie, Huang, and
  Masquelier]{zhu2022training}
Yaoyu Zhu, Zhaofei Yu, Wei Fang, Xiaodong Xie, Tiejun Huang, and Timoth{\'e}e
  Masquelier.
\newblock Training spiking neural networks with event-driven backpropagation.
\newblock In \emph{36th Conference on Neural Information Processing Systems
  (NeurIPS 2022)}, 2022.

\bibitem[Pehle and Pedersen(2021)]{norse2021}
Christian Pehle and Jens~Egholm Pedersen.
\newblock {Norse - A deep learning library for spiking neural networks},
  January 2021.
\newblock URL \url{https://doi.org/10.5281/zenodo.4422025}.
\newblock Documentation: https://norse.ai/docs/.

\bibitem[Fang et~al.(2023)Fang, Chen, Ding, Yu, Masquelier, Chen, Huang, Zhou,
  Li, and Tian]{SpikingJelly}
Wei Fang, Yanqi Chen, Jianhao Ding, Zhaofei Yu, Timoth{\'e}e Masquelier, Ding
  Chen, Liwei Huang, Huihui Zhou, Guoqi Li, and Yonghong Tian.
\newblock Spikingjelly: An open-source machine learning infrastructure platform
  for spike-based intelligence.
\newblock \emph{Science Advances}, 9\penalty0 (40):\penalty0 eadi1480, 2023.

\bibitem[Orchard et~al.(2015)Orchard, Jayawant, Cohen, and
  Thakor]{orchard2015converting}
Garrick Orchard, Ajinkya Jayawant, Gregory~K Cohen, and Nitish Thakor.
\newblock Converting static image datasets to spiking neuromorphic datasets
  using saccades.
\newblock \emph{Frontiers in Neuroscience}, 9:\penalty0 437, 2015.

\bibitem[Cramer et~al.(2020)Cramer, Stradmann, Schemmel, and
  Zenke]{cramer2020heidelberg}
Benjamin Cramer, Yannik Stradmann, Johannes Schemmel, and Friedemann Zenke.
\newblock The heidelberg spiking data sets for the systematic evaluation of
  spiking neural networks.
\newblock \emph{IEEE Transactions on Neural Networks and Learning Systems},
  2020.

\bibitem[Shrestha and Orchard(2018)]{shrestha2018slayer}
Sumit~B Shrestha and Garrick Orchard.
\newblock Slayer: Spike layer error reassignment in time.
\newblock \emph{Advances in neural information processing systems}, 31, 2018.

\bibitem[Zenke and Vogels(2021)]{zenke2021remarkable}
Friedemann Zenke and Tim~P Vogels.
\newblock The remarkable robustness of surrogate gradient learning for
  instilling complex function in spiking neural networks.
\newblock \emph{Neural Computation}, 33\penalty0 (4):\penalty0 899--925, 2021.

\bibitem[Lein et~al.(2007)Lein, Hawrylycz, Ao, Ayres, Bensinger, Bernard, Boe,
  Boguski, Brockway, Byrnes, et~al.]{lein2007genome}
Ed~S Lein, Michael~J Hawrylycz, Nancy Ao, Mikael Ayres, Amy Bensinger, Amy
  Bernard, Andrew~F Boe, Mark~S Boguski, Kevin~S Brockway, Emi~J Byrnes, et~al.
\newblock Genome-wide atlas of gene expression in the adult mouse brain.
\newblock \emph{Nature}, 445\penalty0 (7124):\penalty0 168--176, 2007.

\bibitem[Hawrylycz et~al.(2012)Hawrylycz, Lein, Guillozet-Bongaarts, Shen, Ng,
  Miller, Van De~Lagemaat, Smith, Ebbert, Riley,
  et~al.]{hawrylycz2012anatomically}
Michael~J Hawrylycz, Ed~S Lein, Angela~L Guillozet-Bongaarts, Elaine~H Shen,
  Lydia Ng, Jeremy~A Miller, Louie~N Van De~Lagemaat, Kimberly~A Smith, Amanda
  Ebbert, Zackery~L Riley, et~al.
\newblock An anatomically comprehensive atlas of the adult human brain
  transcriptome.
\newblock \emph{Nature}, 489\penalty0 (7416):\penalty0 391--399, 2012.

\bibitem[van Rossum(2001)]{van2001novel}
Mark~CW van Rossum.
\newblock A novel spike distance.
\newblock \emph{Neural Computation}, 13\penalty0 (4):\penalty0 751--763, 2001.

\bibitem[Eshraghian et~al.(2023)Eshraghian, Ward, Neftci, Wang, Lenz, Dwivedi,
  Bennamoun, Jeong, and Lu]{eshraghian2021training}
Jason~K Eshraghian, Max Ward, Emre~O Neftci, Xinxin Wang, Gregor Lenz, Girish
  Dwivedi, Mohammed Bennamoun, Doo~Seok Jeong, and Wei~D Lu.
\newblock Training spiking neural networks using lessons from deep learning.
\newblock \emph{Proceedings of the IEEE}, 2023.

\bibitem[Pozzorini et~al.(2015)Pozzorini, Mensi, Hagens, Naud, Koch, and
  Gerstner]{pozzorini2015automated}
Christian Pozzorini, Skander Mensi, Olivier Hagens, Richard Naud, Christof
  Koch, and Wulfram Gerstner.
\newblock Automated high-throughput characterization of single neurons by means
  of simplified spiking models.
\newblock \emph{PLoS Computational Biology}, 11\penalty0 (6):\penalty0
  e1004275, 2015.

\bibitem[Stuijt et~al.(2021)Stuijt, Sifalakis, Yousefzadeh, and
  Corradi]{stuijt2021mubrain}
Jan Stuijt, Manolis Sifalakis, Amirreza Yousefzadeh, and Federico Corradi.
\newblock $\mu$brain: An event-driven and fully synthesizable architecture for
  spiking neural networks.
\newblock \emph{Frontiers in Neuroscience}, page 538, 2021.

\bibitem[He et~al.(2021)He, Corradi, Shi, Ding, Timmermans, Stuijt, Harpe,
  Ocket, and Liu]{he202128}
Yuming He, Federico Corradi, Chengyao Shi, Ming Ding, Martijn Timmermans, Jan
  Stuijt, Pieter Harpe, Ilja Ocket, and Yao-Hong Liu.
\newblock A 28.2 $\mu$c neuromorphic sensing system featuring snn-based
  near-sensor computation and event-driven body-channel communication for
  insertable cardiac monitoring.
\newblock In \emph{2021 IEEE Asian Solid-State Circuits Conference (A-SSCC)},
  pages 1--3. IEEE, 2021.

\bibitem[Panda et~al.(2020)Panda, Aketi, and Roy]{panda2020toward}
Priyadarshini Panda, Sai~Aparna Aketi, and Kaushik Roy.
\newblock Toward scalable, efficient, and accurate deep spiking neural networks
  with backward residual connections, stochastic softmax, and hybridization.
\newblock \emph{Frontiers in Neuroscience}, 14:\penalty0 653, 2020.

\bibitem[Chetlur et~al.(2014)Chetlur, Woolley, Vandermersch, Cohen, Tran,
  Catanzaro, and Shelhamer]{chetlur2014cudnn}
Sharan Chetlur, Cliff Woolley, Philippe Vandermersch, Jonathan Cohen, John
  Tran, Bryan Catanzaro, and Evan Shelhamer.
\newblock cudnn: Efficient primitives for deep learning.
\newblock \emph{arXiv:1410.0759}, 2014.

\end{thebibliography}

\end{document}


\maketitle

\section{Theoretical proofs}
Proofs for all the propositions in the paper, outlining the mathematical equivalence of our Block model to the standard ALIF SNN.

\begin{proposition}
\label{prop:conv_mem}
Membrane potentials without spike reset are computed as a convolution $\tilde{V}_i[t] = \big(I_i * \tilde{\beta}_i\big)[t]$ between input current $I_i[t]$ and kernel $\tilde{\beta}_i[t]=(1-\beta_i)\beta_i^t$ with the initial membrane potential encoded as $I_i[0]=\frac{V_i[0]}{1-\beta_i}$.
\begin{proof}
We proceed our proof in two steps. In step 1, we unroll the discretized LIF difference equation (without reset) in time and in step 2, we show how this is equivalent to the proposed convolution. 
*
Step 1, we prove the equivalence between the following equations
\begin{align}
	\tilde{V}_i[t] &= \beta_i \tilde{V}_i[t-1] + (1-\beta_i)I_i[t] \label{eq:disc_lif} \\
	\tilde{V}_i[t] &= \beta_i^{t} \tilde{V}_i[0] + (1-\beta_i) \sum_{j=1}^{t} \beta_i^{t-j} I_i[j] \label{eq:disc_lif_rollout}
\end{align}
We proceed by induction. For $t=1$ in Equation 2 we obtain
\begin{equation}
	\begin{split}
		\tilde{V}_i[1] &= \beta_i^{1} \tilde{V}_i[0] + (1-\beta_i) \sum_{j=1}^{1} \beta_i^{1-j} I_i[j]\\
			&= \beta_i^{1} \tilde{V}_i[0] + (1-\beta_i) I_i[1]
	\end{split}
\end{equation}
Hence the relation holds true for the base case $t=1$. Assume the relation holds true for $t=k\geq1$, then for $t=k+1$ we derive
\begin{equation}
	\begin{split}
		\tilde{V}_i[k+1] &= \beta_i \tilde{V}_i[k] + (1 - \beta_i) I_i[k+1]\\
		&= \beta_i \Big( \beta_i^{k} \tilde{V}_i[0] + (1-\beta_i) \sum_{j=1}^{k} \beta_i^{k-j} I_i[j] \Big) + (1 - \beta_i) I_i[k+1]\\
		&= \beta_i^{k+1} \tilde{V}_i[0] + (1-\beta_i) \sum_{j=1}^{k} \beta_i^{(k+1)-j} I_i[j] + (1-\beta_i)I_i[k+1]\\
		&= \beta_i^{k+1} \tilde{V}_i[0] + (1-\beta_i) \sum_{j=1}^{k+1} \beta_i^{(k+1)-j} I_i[j]
	\end{split}
\end{equation}
This implies equivalence between Equations 1 and 2 for $t=k+1$ assuming equivalence between Equations 1 and 2 holds true for $t=k$. By the principle of induction, equivalence is established given that both the base case and inductive step hold true.

Step 2, as per the proposition, we have
\begin{equation}
	\begin{split}
	\tilde{V}_i[t] &= \big(I_i * \tilde{\beta}_i\big)[t]\\
	&= \sum_{j=0}^{t} \tilde{\beta}_i[t-j] I_i[j]\\
	&= (1-\beta_i)\sum_{j=0}^{t} \beta_i^{t-j} I_i[j]\\
	&= (1-\beta_i)\beta_i^tI_i[0] + (1-\beta_i)\sum_{j=1}^{t} \beta_i^{t-j} I_i[j]\\
	&= \beta_i^t\tilde{V}_i[0] + (1-\beta_i)\sum_{j=1}^{t} \beta_i^{t-j} I_i[j]\\
	\end{split}
\end{equation}
This is identical to Equation \ref{eq:disc_lif_rollout} and (by step 1) identical to Equation \ref{eq:disc_lif}. Thus, the proposed convolution in the Proposition computes the membrane potentials without reset.
\end{proof}	 
\end{proposition}

\begin{proposition}
\label{prop:phi}
Function $\phi(\tilde{S}_i)[t]=\sum_{k=1}^{t}\tilde{S}_i[k](t-k+1)$ acting on $\tilde{S}_i \in \{0, 1\}^T$ contains at most one element equal to one $\phi(\tilde{S}_i)[t]=1$ for smallest $t$ satisfying $\tilde{S}_i[t]=1$ (if such $t$ exists).
\begin{proof}
Firstly, if $\tilde{S}^{(l)}_i[t]=0$ for all $t \in [1, T]$ then $\phi(\tilde{S}^{(l)}_i)[t]=0$ for all $t \in [1, T]$ (follows from substitution). Secondly, if $\tilde{S}^{(l)}_i[t_1]=1$ for smallest $t_1 \in [1, T]$ then $\phi(\tilde{S}^{(l)}_i)[t_1]=1$ (follows from substitution) and there can exist no $t_2>t_1$ such that $\phi(\tilde{S}^{(l)}_i)[t_2]=1$ as
\begin{equation}
	\begin{split}
		\phi(\tilde{S}^{(l)}_i)[t+1]&=\sum_{k=1}^{t+1}\tilde{S}^{(l)}_i[k]\big((t+1)-k+1\big) \\
		&=\sum_{k=1}^{t}\tilde{S}^{(l)}_i[k]\big((t+1)-k+1\big)+\tilde{S}^{(l)}_i[t+1]\\
		&=\sum_{k=1}^{t}\tilde{S}^{(l)}_i[k](t-k+1)+\sum_{k=1}^{t}\tilde{S}^{(l)}_i[k]+\tilde{S}^{(l)}_i[t+1]\\
		&=\phi(\tilde{S}^{(l)}_i)[t]+\sum_{k=1}^{t+1}\tilde{S}^{(l)}_i[k]
	\end{split}
\end{equation}
Thus $\phi(\tilde{S}^{(l)}_i)[t_2]>\phi(\tilde{S}^{(l)}_i)[t_1]$ for all $t_2>t_1$ as $\sum_{k=1}^{t_2}\tilde{S}^{(l)}_i[k]\geq\sum_{k=1}^{t_1}\tilde{S}^{(l)}_i[k]=1>0$.
\end{proof}
\end{proposition}

\begin{proposition}
\label{prop:blocks_input_current}
The input current $I_{i, n+1}[t]$ of neuron $i$ simulated in Block $n+1$ (of length $T_R$) is defined as follows, and enforces an absolute refractory period of length $T_R$ and a monosynaptic transmission latency of $D=T_R$.
\begin{equation}
\nonumber
I_{i, n+1}[t] = \Big( b_{i} + \underbrace{\sum_{j=1}^{N^\text{in}} W_{ij} S_{j, n+1}[t]}_\text{Feedforward current} + \underbrace{\sum_{j=1}^{N^\text{out}} W^{\text{rec}}_{ij} S_{j, n}[t]}_\text{Recurrent current} \Big) \underbrace{\mathbb{1}_{\displaystyle z_{i, n}[t] \geq \max_t S_{i, n}[t]}\vphantom{\sum_{j=1}^{N^{(l)}}}}_\text{ARP mask}
\end{equation}
\begin{proof}
We proceed by showing that 1. the transmission latency is the same as in the standard model and 2. the ARP mask enforces an ARP of identical length to the standard model. 

1. Identical transmission latency: The relation between the time step $1 \leq t \leq T$ and the time step $1 \leq t_b \leq T_R$ in Block $n \geq 1$ can be expressed as:
\begin{equation}
	t = (n-1) T_R + t_b
\end{equation}
As we assumed the transmission latency $D=T_R$ to be equal to the ARP length, we have
\begin{equation}
	t - T_R = ((n-1)-1) T_R + t_b
\end{equation}
and hence the input current $I_{i, n}[t]$ to Block $n$ at time $t$ (\textit{i.e.} time step $t_b$ in the Block) is computed from the output spikes from the prior Block $n-1$ at the same Block time step $t_b$.

2. Identical ARP length: Case one, if neuron $i$ emitted no spikes during Block $n$, then neuron $i$ should receive input current at every time step in Block $n+1$, which the ARP mask permits (as $z_{i, n}[t] \geq \max_t S_{i, n}[t] = 0$ for all $t$). Case two, if neuron $i$ spiked during Block $n$, then the ARP mask should appropriately mask out the input current to Block $n+1$ (to enforce an ARP of length $T_R$). The number of elements in $z_{i, n}$ which are smaller than one and equal to or larger than one is equal to $T_R$ (can be shown by proof by contradiction):
\begin{equation}
	\underbrace{\sum \mathbb{1}_{\displaystyle z_{i, n}[t] < 1}}_\text{Number of elements masked in Block $n + 1$} + \underbrace{\sum \mathbb{1}_{\displaystyle z_{i, n}[t] \geq 1}}_\text{Number of time steps from the spike onwards in Block $n$} = T_R
\end{equation}

The construction of the Block ensures that at most one spike is emitted within a Block (where $z_{i,n}[t]=1$) and no further spikes are emitted thereafter (where $z_{i,n}[t]>1$). The ARP mask ensures that no spikes are emitted in the next Block for the time steps where $z_{i,n}[t]<1$ (as it only permits input current to flow into the Block for time steps where $z_{i,n}[t] \geq 1$). Thus, the mask ensures enforces the correct ARP length of $T_R$ steps.
\end{proof}
\end{proposition}

\begin{proposition}
\label{prop:blocks_membrane}
The initial membrane potential $V_{i, n+1}[0]$ of neuron $i$ simulated in Block $n+1$ (of length $T_R$) is equal to the last membrane potential in Block $n$ if no spike occurred and zero otherwise.
\begin{equation}
\nonumber
V_{i, n+1}[0]=\begin{cases}
    V_{i, n}[T_R], & \text{if } \max_t S_{i, n}[t] = 0 \\
    0, & \text{otherwise}
\end{cases}
\end{equation} 
\begin{proof}
Two cases are distinguished for correctly evolving the membrane potential of a neuron over time. Case one, if no spike occurred in neuron $i$ in Block $n$ (\textit{i.e.} $ \max_t S_{i, n}[t] = 0$), then the initial membrane potential of neuron $i$ in Block $n+1$ is equal to the final membrane potential value of neuron $i$ in Block $n$. Otherwise, case two, the initial membrane potential is set to zero (as no state needs to be transferred as the neuron is in an absolute refractory state).
\end{proof}
\end{proposition}

\begin{proposition}
\label{prop:blocks_adaptive_thresh}
The adaptive firing threshold $\theta_{i, n+1}[t]$ of neuron $i$ simulated in Block $n+1$ (of length $T_R$) is constructed from the initial adaptive parameter $a_{i, n+1}[0]$, which is equal to its last value in the previous Block if no spike occurred, and otherwise equal to an expression which accounts for the effect of the spike on the adaptive threshold.
\begin{align}
\theta_{i, n+1}[t] &= 1 + d_i p_i^{t} a_{i, n+1}[0] \nonumber\\
a_{i, n+1}[0]&=\begin{cases}
    a_{i, n}[T_R], & \text{if } \max_t {S_{i, n}[t]} = 0 \nonumber\\
    p_{i}^m (a_s + p_i^{-1}) & \text{otherwise}
\end{cases}\\
m = \sum_{k}^{T_R} &\mathbb{1}_{\displaystyle z_{i, n}[k]>1}, \quad a_s = \sum_{k}^{T_R} a_{i, n}[k] S_{i, n}[k] \nonumber
\end{align}
\begin{proof}
We proceed our proof in two steps. In step 1, we show how the dynamic firing threshold of neuron $i$ in Block $n+1$ can be computed using initial adaptive parameter $a_{i, n+1}[0]$; and in step 2, we show how this initial adaptive parameter is derived.

Step 1, we prove the equivalence between the Block adaptive threshold (Equation \ref{eq:blocks_adapt}) and the standard adaptive firing threshold (Equations \ref{eq:standard_adapt1} and \ref{eq:standard_adapt2}).
\begin{align}
\theta_{i, n+1}[t] &= 1 + d_i p_i^{t} a_{i, n+1}[0] \label{eq:blocks_adapt}\\
	\theta_{i, n+1}[t] &= 1 + d_{i} a_{i, n+1}[t] \label{eq:standard_adapt1} \\
	a_{i, n+1}[t] &= p_{i} a_{i, n+1}[t-1] + S_{i, n+1}[t-1]
 \label{eq:standard_adapt2}
\end{align}
The spike term in Equation \ref{eq:standard_adapt2} can be dropped, as only the spike occurrence in Block $n$ (and not Block $n+1$) can affect the firing threshold in Block $n+1$ (due to the single spike constraint). Thus, we rewrite Equation \ref{eq:standard_adapt1} as:
\begin{align}
\theta_{i, n+1}[t] &= 1 + d_{i} a_{i, n+1}[t]\nonumber\\
&= 1 + d_{i} p_i a_{i, n+1}[t-1]\nonumber \\
&= 1 + d_{i} p_i^2 a_{i, n+1}[t-2]\nonumber \\
& \cdots\nonumber \\
&= 1 + d_{i} p_i^t a_{i, n+1}[0]\nonumber
\end{align}

Step 2, to derive the initial adaptive parameter $a_{i, n+1}[0]$, we distinguish two cases. Case one, neuron $i$ did not spike in Block $n$, in which case the initial adaptive parameter is set to the final adaptive parameter $a_{i, n}[T_R]$ in Block $n$. Case two, neuron $i$ did spike in Block $n$, and we need to account for the affect of this on the firing threshold in Block $n+1$. If the spike occurred at Block time step $1 \leq T_R - m \leq T_R$ for $0 \leq m < T_R$ we have
\begin{align}
a_{i, n}[T_R] &= p_i a_{i, n}[T_R - 1]\nonumber\\
&= p_i^2 a_{i, n}[T_R - 2]\nonumber\\
& \cdots\nonumber\\
&= p_i^{m-1} a_{i, n}[T_R - (m-1)]\nonumber\\
&= p_i^{m-1} \Big(p_i a_{i, n}[T_R - m] + 1 \Big)\nonumber\\
&= p_i^{m} \Big(a_{i, n}[T_R - m] + p_i^{-1} \Big)\nonumber\\
&= p_i^{m} \Big(a_s + p_i^{-1} \Big)
\end{align}
with $a_s=\sum_{k}^{T_R} a_{i, n}[k] S_{i, n}[k]=a_{i, n}[T_R - m]$ (as $S_{i, n}[k]$ is one for $k = T_R - m$ and zero otherwise). Lastly, $m = \sum_{k}^{T_R} \mathbb{1}_{\displaystyle z_{i, n}[k]>1}$, as $z_{i, n}[k]>1$ at ever Block time step $k > T_R - m$ (see Proposition \ref{prop:phi} proof showing $z_{i, n}$ to be a strictly increasing sequence if a spike occurred in Block $n$) and thus $\sum_{k}^{T_R} \mathbb{1}_{\displaystyle z_{i, n}[k]>1}=\sum_{k=T_R-m+1}^{T_R} 1=T_R-(T_R-m)=m$.
\end{proof}
\end{proposition}

\section{Experimental details}

All models were implemented using PyTorch \citep{paszke2017automatic} (although nothing prohibits the use of other auto differentiation frameworks \cite{paszke2017automatic, jax2018github}). The speedup benchmarks and neural fits were done on an NVIDIA GeForce RTX 3090, and the training on the spiking classification datasets was done on an NVIDIA GeForce GTX 1080 Ti. Following details apply to both our Block model and standard SNN model which we used as a control.

\subsection{Speed benchmarks}

\paragraph{Synthetic spike dataset}
We generated binary input spike tensors of shape $B \times N \times T$ ($B$ being the batch size, $N$ the number of input neurons and $T$ the number of time steps). For every batch dimension $b$ a firing rate $r_b \sim \mathbf{U}(u_\text{min}, u_\text{max})$ was uniformly sampled (with $u_\text{min}=0$Hz and $u_\text{max}=200$Hz – Assuming $1$ time step $=1$ms), from which a random binary spike matrix of shape $N \times T$ was constructed as a homogenous Poisson process,  such that every input neuron in this matrix had a firing rate of $r_b$Hz.

\subsection{Supervised learning}
\paragraph{Datasets} We tested our model (and control) on two common spike classification datasets, the Neuromophic-MNIST (N-MNIST) \cite{orchard2015converting} and Spiking Heidelberg Digits (SHD) \cite{cramer2020heidelberg} dataset (both released under the Creative Commons Attribution 4.0 International License). The N-MNIST dataset is the classical MNIST dataset mapped onto a spike code using a neuromorphic vision sensor and the SHD dataset comprises spoken digit waveforms converted into spikes using a model of the auditory bushy neurons in the cochlear nucleus. 

\paragraph{Weight initialisation} All network connectivity weights were sampled from a uniform distribution $\mathbf{U}(-\sqrt{N^{-1}}, \sqrt{N^{-1}})$ with $N$ number of afferent connections. All biases were initialised as $0$. The hidden neurons were initialised with a membrane time constant of $20$ms (\textit{i.e.} $\beta_i^{(l)}=\exp(\frac{-DT}{20})$), an adaptive time constant of $150$ms (\textit{i.e.} $p_i^{(l)}=\exp(\frac{-DT}{150})$) and adaptive parameter of $d_i^{(l)}=1.8$. The readout neurons were initialised with a membrane time constant of $20$ms.

\paragraph{Clamping time constants} To enforce correct neuron dynamics, we clamped the values of $\beta_i^{(l)}$ into the range $[0.01, 0.99]$ and the values of $p_i^{(l)}$ into the range $[0.0, 0.999]$
\begin{equation}
\beta_i^{(l)}=\begin{cases}
    0.99, & \text{if } \beta_i^{(l)} > 0.99 \\
    0.01, & \text{if } \beta_i^{(l)} < 0.01 \\
\end{cases}
\end{equation}
\begin{equation}
p_i^{(l)}=\begin{cases}
    0.999, & \text{if } p_i^{(l)} > 0.999 \\
    0.0, & \text{if } p_i^{(l)} < 0.0 \\
\end{cases}
\end{equation}

\paragraph{Readout neurons} Every network had an output layer of readout neurons (containing the same number of neurons as the number of classes within the dataset trained on), where we removed the spike and reset mechanism (as done in \cite{zenke2021remarkable}). The output of the readout neuron $c$ in response to input sample $b$ was taken to be the summated membrane potential over time  $o_{b, c} = \sum_t V_{b, c}^{L}[t]$ ($L$ being the readout layer).

\paragraph{Supervised training loss} We trained all networks to minimise a cross-entropy loss (with $B$ and $C$ being the number of batch samples and dataset classes respectively)
\begin{equation}
	\mathcal{L} = -\frac{1}{B} \sum_{b=1}^{B} \sum_{c=1}^{C} y_{bc} \log(\hat{y}_{bc})
\end{equation}
Variable $y_{bc} \in \{0,1\}^C$ is the one hot target vector and $\hat{y}_{bc} $ are the network prediction probabilities, which were obtained by passing the readout neuron outputs $o_{bc}$ through the softmax function.
\begin{equation}
	 \hat{y}_{bc} = \frac{\exp(o_{bc})}{\sum_{k=1}^{C} \exp(o_{bk})}
\end{equation}

\paragraph{Surrogate gradient} We tested training on the SHD dataset using three different surrogate gradient functions, including the multi-Gaussian \cite{yin2021accurateMG}, fast sigmoid \citep{zenke2018superspike} and the boxcar \cite{kaiser2020synaptic, bittar2022surrogate} function - all which have shown to perform well in training SNNs.
\begin{align}
\frac{\partial S_i^{(l)}[t]}{\partial V_i^{(l)}[t]}&=1.15 \mathcal{N}(V_i^{(l)}[t] \mid 0, 0.5^2) - 0.15 \mathcal{N}(V_i^{(l)}[t] \mid 3, 3^2) \\&- 0.15 \mathcal{N}(V_i^{(l)}[t] \mid -3, 3^2) \quad \textrm{(multi-Gaussian)} \\
\frac{\partial S_i^{(l)}[t]}{\partial V_i^{(l)}[t]}&=(10 |V_i^{(l)}[t]|+1)^{-2} \quad \textrm{(fast sigmoid)} \\
\frac{\partial S_i^{(l)}[t]}{\partial V_i^{(l)}[t]}&=\begin{cases}
    0.5, & \text{if } |V_i^{(l)}[t] - \theta_i^{(l)}| \leq 0.5 \\
    0, & \text{otherwise}
\end{cases} \quad \textrm{(boxcar)} \\
\end{align}
We chose the multi-Gaussian function as this obtained the best performance (Supplementary Figure \ref{fig:sg_supp}a).

\paragraph{Detaching gradients} Detaching gradients from flowing through the spike reset and recurrent connections has been shown to improve classification accuracies when training SNNs \cite{zenke2021remarkable}. We thus explored 1. allowing gradients to flow through all elements within the computational graph (\textit{i.e.} attached) and 2. detaching the surrogate gradient from all elements within the computational graph, except for the feedforward connections to other neurons (\textit{i.e.} detached). We found that detaching improved test accuracies across all the tried surrogate gradient functions on the SHD dataset (Supplementary Figure \ref{fig:sg_supp}b).

\paragraph{Training procedure} We used the Adam optimiser (with default parameters) \citep{kingma2014adam} for all training, starting with an initial learning rate of $0.001$, which was decayed by a factor of $10$ every time the number of epochs reached a new milestone. Model weights were saved if the training error at the end of each epoch was lowered. The hyperparameters are found in Supplementary Table \ref{table:hyperparams}.

\paragraph{Additional results} Forward (\textit{i.e.} inference using the network) speedup can be found in Supplementary Figure \ref{fig:speed_supp}. Here we used the same simulation setup as the one we used for benchmarking the training speedups. We also compared the forward pass of our model to other publicly available SNN implementations, the Norse library \cite{norse2021} and the Spiking Jelly library \cite{SpikingJelly}, and found our method to run considerably faster (Supplementary Table \ref{table:library_comparison}). We benchmarked a simple one layer SNN network of $100$ units on the forward pass, over $1000$ time steps with $200$ input units (with a batch size of $128$). As expected, we found the Norse ($0.30$s) and SpikingJelly ($0.18$s) implementations took a similar time to our standard SNN implementation ($0.33$s) with the SpikingJelly implementation running slightly faster (as they are all governed by the same sequential time complexity). In contrast, our Blocks model (using a $50$ steps ARP) ran over an order of magnitude faster ($0.016$s) than the other implementations.

\subsection{Neural fits}

\paragraph{Dataset preprocessing}
We fitted the model to \textit{in vitro} whole-cell patch clamp electrophysiological recordings from neurons in layer 4 in mouse primary visual cortex, provided by the Allen institute \cite{lein2007genome, hawrylycz2012anatomically}. These neurons were injected with a varying current at different amplitudes. For each neuron, the training and test datasets were those stipulated by the Allen Institute ($50\%$ for training and $50\%$ for testing). We only fitted and reported our results on neurons which had four repeats for each unique current injection. We processed the input data by 1. removing all the long periods in the recordings during which no current was injected, 2. resampling the data to have a DT$=0.1$ms, and lastly 3. normalising the data by subtracting the mean and by dividing by the standard deviation of the training dataset. We used the spike times provided by the Allen Institute for fitting.

\paragraph{Weight initialisation} 
The neuron input weight was initialised to a constant value of $\frac{s}{100}$ for scale value $s$, ensuring that the input weight was correctly rescaled for different simulation resolutions (\textit{e.g.} $s=1$ corresponds to simulating with DT$=0.1$ms and $s=2$ corresponds to simulating with DT$=0.2$ms). The bias was initialized to $0$. The membrane time constant was initialized to $20$ms, the adaptive time constant was initialized to $100$ms, and the adaptive parameter was initialized to $d=\frac{0.1}{s}$.

\paragraph{Supervised training loss} Each ALIF neuron model was optimised to produce the same spike times of the real neuron being fit to. This was achieved by minimising the van Rossum distance $D_R$ \cite{van2001novel} between the predicted model spike train $x$ and real neuron spike train $y$, defined as
\begin{equation}
	D_R(\tau_R) = \sqrt{\frac{1}{\tau_R}\int_0^T \big(k * x(t) - k * y(t)\big)^2dt}
\end{equation}
with exponential kernel $k=H(t)\exp(\frac{-t}{\tau_R})$ and $H$ being the heaviside function. We used $\tau_R=100$ms for all our reported results.

\paragraph{Training procedure} We used the Adam optimiser (with default parameters) \citep{kingma2014adam} for every neuron fitted, with a learning rate of $0.0001$. Training was carried out over $200$ epochs in full batch mode (\textit{i.e.} estimating gradients from the entire training dataset). Model parameters were saved whenever the training score improved, and training was halted if there was no improvement over the last five epochs.

\paragraph{Explained temporal variance} We used the explained temporal variance (ETV) metric to assess the goodness-of-fit of the fitted models to the neural data (used on the Allen Institute website). The metric has a value of zero when the model predicts at chance and a value of one for a perfect fit to the data, and is defined as:
\begin{equation}
	\text{ETV} = \frac{\text{ETV}_{\text{raw}}}{\text{ETV}_{\text{max}}}
\end{equation}
$\text{ETV}_{\text{raw}}$ measures the pairwise explained variance of the data with the model and $\text{ETV}_{\text{max}}$ measures the upper limit on how well the model can perform (\textit{i.e.} how much of the neuron's response variability from repeat to repeat can be accounted for).
\begin{align}
\text{ETV}_{\text{raw}} &= \sum_r \frac{\Var(g * x) + \Var(g * y_r) - \Var(g * x - g * y_r)}{\Var(g * x) + \Var(g * y_r)} \\
 \text{ETV}_{\text{max}} &= \sum_r \frac{\Var(g * \bar{y}) + \Var(g * y_r) - \Var(g * \bar{y} - g * y_r)}{\Var(g * \bar{y}) + \Var(g * y_r)}
\end{align}
Here $x$ is the predicted model spike train, $y_r$ is recorded neuron spike train for repeat $r$, $g$ is a Gaussian kernel (with mean $\mu=0$ and standard deviation  $\sigma=150$ms) and $\bar{y}[t]=\frac{1}{R} \sum_r^R (g * y_r)[t]$ is the mean recorded (and smoothed) neuron response. Convolving the spike trains with the Gaussian kernel converts them into peristimulus time histograms.

\paragraph{Additional results} Zoomed-in plots of the neural traces can be found in Supplementary Figure \ref{fig:trace_supp}, and the membrane time constant distribution of the fitted V1 neurons can be found in Supplementary Figure \ref{fig:mem_dist}.

\bibliography{bibfile}

\begin{table}[h]\label{table:hyperparameters}
\caption{\textbf{Dataset and corresponding training parameters.}}
\label{table:hyperparams}
\begin{center}
\begin{tabular}{ccc}
\hline
&\multicolumn{1}{c}{\bf N-MNIST} &\multicolumn{1}{c}{\bf SHD}\\
\hline
\multicolumn{1}{c}{\multirow{1}{*}{Dataset (train/test)}} &$60\text{k}/10\text{k}$&$8156/2264$ \\
\multicolumn{1}{c}{\multirow{1}{*}{Input neurons}} &$1156$ &$700$ \\
\multicolumn{1}{c}{\multirow{1}{*}{Dataset classes}} &$10$ &$20$ \\
\multicolumn{1}{c}{\multirow{1}{*}{Epochs}} &$30$ &$40$ \\
\multicolumn{1}{c}{\multirow{1}{*}{Batch size $B$}} &$64$ &$64$ \\
\multicolumn{1}{c}{\multirow{1}{*}{Time steps $T$}} &$300$ &$600$ \\
\multicolumn{1}{c}{\multirow{1}{*}{Time resolution $\Delta t$} (ms)} &$1$ &$2$ \\
\multicolumn{1}{c}{\multirow{1}{*}{LR decay epoch milestones}} &$N/A$ &$(15, 30)$ \\
\hline
\end{tabular}
\end{center}
\end{table} 

\begin{table}[h]
\caption{\textbf{Comparison to other SNN libraries}. Forward pass simulation duration of our model compared to other publicly available SNN implementations.}
\label{table:library_comparison}
\begin{center}
\begin{tabular}{cc}
\hline
&\multicolumn{1}{c}{\bf Duration (s)}\\
\hline
\multicolumn{1}{c}{\multirow{1}{*}{SpikingJelly \cite{norse2021}}} &$0.18$s \\
\multicolumn{1}{c}{\multirow{1}{*}{Norse \cite{SpikingJelly}}} &$0.30$s \\
\multicolumn{1}{c}{\multirow{1}{*}{Standard SNN (our implementation)}} &$0.33$s \\
\multicolumn{1}{c}{\multirow{1}{*}{Block SNN (our model)}} &$0.016$s \\
\hline
\end{tabular}
\end{center}
\end{table}

\begin{figure}[h]
    \includegraphics[width=1\linewidth]{figures/supplementary/sg_supp.pdf}
	\centering
	\caption{\textbf{Surrogate gradient search.} \textbf{a.} Accuracy on the SHD dataset using our model for different surrogate gradient functions. \textbf{b.} Accuracy on the SHD dataset using our model when attaching the surrogate gradient to all elements within the computational graph vs detaching it from everything besides the connections to efferent neurons. Bars plot the mean and standard error  over three runs.}
 	\label{fig:sg_supp}
\end{figure}
\begin{figure}[h]
    \includegraphics[width=1\linewidth]{figures/supplementary/speed_supp.pdf}
	\centering
	\caption{\textbf{Forward speedup of our model.} \textbf{a.} Forward speedup of our accelerated ALIF model compared to the standard ALIF model for different simulation lengths $T$, ARP time steps and batch sizes. \textbf{b.} Forward speedup over different number of layers and hidden units (with ARP time steps$=40$, $T=1024$ and batch$=64$; bars plot the mean and standard error over ten runs). Assuming DT$=0.1$ms, then $10$ time steps $=1$ms.}
 	\label{fig:speed_supp}
\end{figure}

\begin{figure}[h]
    \includegraphics[width=1\linewidth]{figures/supplementary/zoom5b.pdf}
	\centering
	\caption{\textbf{Zoomed-in versions of the neural traces from Figure 5b.}}
 	\label{fig:trace_supp}
\end{figure}

\begin{figure}[h]
    \includegraphics[width=1\linewidth]{figures/supplementary/membrane_dist.pdf}
	\centering
	\caption{\textbf{Membrane time constant distribution of the neurons fitted in Figure 5.}}
	\label{fig:mem_dist}
\end{figure}